\documentclass[10pt,twocolumn,letterpaper]{article}

 \usepackage{cvpr}              
\usepackage[accsupp]{axessibility} 
\usepackage[dvipsnames]{xcolor}

\definecolor{cvprblue}{rgb}{0.21,0.49,0.74}
\usepackage[pagebackref,breaklinks,colorlinks,citecolor=cvprblue]{hyperref}
\usepackage{graphicx}
\usepackage{amsmath}
\usepackage{amssymb}
\usepackage{subfloat}
\usepackage{multirow}
\usepackage{enumitem}
\usepackage[title]{appendix}
\usepackage[ruled,linesnumbered]{algorithm2e}
\usepackage{color}

\def\paperID{} 
\def\confName{CVPR}
\def\confYear{2024}

\title{MP-PolarMask: A Faster and Finer Instance Segmentation for Concave Images}

\author{
Ke-Lei Wang$^1$, 
Pin-Hsuan Chou$^1$, 
Young-Ching Chou$^1$,
Chia-Jen Liu$^2$, \and
Cheng-Kuan Lin$^1$, 
Yu-Chee Tseng$^1$\\
$^1$Department of Computer Science\\
$^2$Institute of Emergency and Critical Care Medicine\\
National Yang Ming Chiao Tung University, Hsinchu, Taiwan\\
{\tt\small
kevinfat880414@gmail.com, \{sherrychou.sc09, melody.c, chiajenliu\}@nycu.edu.tw}\\
{\tt\small\{cklin, yctseng\}@cs.nycu.edu.tw}
}

\begin{document}
\maketitle
\begin{abstract}
While there are a lot of models for instance segmentation, PolarMask stands out as a unique one that represents an object by a Polar coordinate system. With an anchor-box-free design and a single-stage framework that conducts detection and segmentation at one time, PolarMask is proved to be able to balance efficiency and accuracy. Hence, it can be easily connected with other downstream real-time applications. In this work, we observe that there are two deficiencies associated with PolarMask: (i) inability of representing concave objects and (ii) inefficiency in using ray regression. We propose MP-PolarMask (Multi-Point PolarMask) by taking advantage of multiple Polar systems. The main idea is to extend from one main Polar system to four auxiliary Polar systems, thus capable of representing more complicated convex-and-concave-mixed shapes. We validate MP-PolarMask on both general objects and food objects of the COCO dataset, and the results demonstrate significant improvement of $13.69\%$ in AP$_L$ and $7.23\%$ in $AP$ over PolarMask with 36 rays.
\end{abstract}

\section{Introduction}
\label{sec:intro}

Computer vision techniques have been widely used in various areas \cite{video-retrieval-CVPR, video-retrieval-PR, boundary, 5G-CV}. Food science is gaining popularity with the growing emphasis on health. In particular, food segmentation offers valuable insights for calories estimation \cite{poply2021refined} and food waste statics \cite{rahman2024kitchen}. Food segmentation presents a substantial challenge problem due to the diverse nature of its appearance and intra-class variations \cite{aslan2020benchmarking}. Therefore, our focus is directed towards convex-and-concave-mixed images, like those food images in the COCO dataset \cite{10.1007/978-3-319-10602-1_48}.

Food segmentation can be considered as one of the applications of instance segmentation. Instance segmentation stands as a crucial subfield within computer vision \cite{9878483, 10.1007/978-3-319-10584-0_20, 10.5555/2969442.2969462}. Its primary purpose is to address the challenge of identifying specific objects or targets within an entire image while providing crucial information such as the target's category, precise location, and accurate segmentation boundaries. It can be seen as a combination of semantic segmentation and object detection.
Object detection systems roughly localize multiple objects using bounding boxes, while semantic segmentation frameworks assign category information to each pixel for a class. In contrast, instance segmentation takes a step further by labeling each pixel with a specific instance, rather than just a particular class.
This enhanced approach allows for more meaningful and detailed inferences on an image, which finds practical applications in various domains, enabling tasks such as object localization, recognition, and comprehensive scene understanding.

Instance segmentation primarily revolves around two frameworks: two-stage frameworks and one-stage frameworks. The two-stage instance segmentation can be implemented using two distinct approaches: the bottom-up method \cite{8237640, Brabandere_2017_CVPR_Workshops}, which relies on semantic segmentation, and the top-down method, which is based on detection \cite{10.5555/2969442.2969462, 10.1007/978-3-319-46448-0_5, he2018mask, 8099955, Huang_2019_CVPR}. 
In platforms with abundant computing resources, the two-stage frameworks tend to achieve higher accuracy,  but spend much time on heavy computation, thus limiting their applications in real-time tasks. To resolve this issue, there are some upcoming approaches that employ a one-stage pipeline for both object detection and instance segmentation. One-stage frameworks often have a simpler structure, incorporate a lightweight backbone, deal with fewer candidate areas, and employ fully convolutional detection networks, thus generally running faster than two-stage approaches \cite{bolya2019yolact, 9159935, 10.1007/978-3-030-58523-5_38, 9746109, cheng2022sparse, tian2019fcos}.

PolarMask \cite{9157078} is an anchor-free and one-stage instance segmentation method that is characterized by its simplicity in concept and fully convolutional nature. The key advantage lies in its seamless integration into most off-the-shelf detection methods, enabling a high level of adaptability and practicality for diverse applications. The method yields good results when dealing with those convex-shaped objects. However, it encounters challenges when dealing with more complex objects, especially those concave-shaped ones.

\begin{figure} [t]
    \centering 
    \subfloat[Mask points of Polarmask]
    {\includegraphics[height=1.1in]{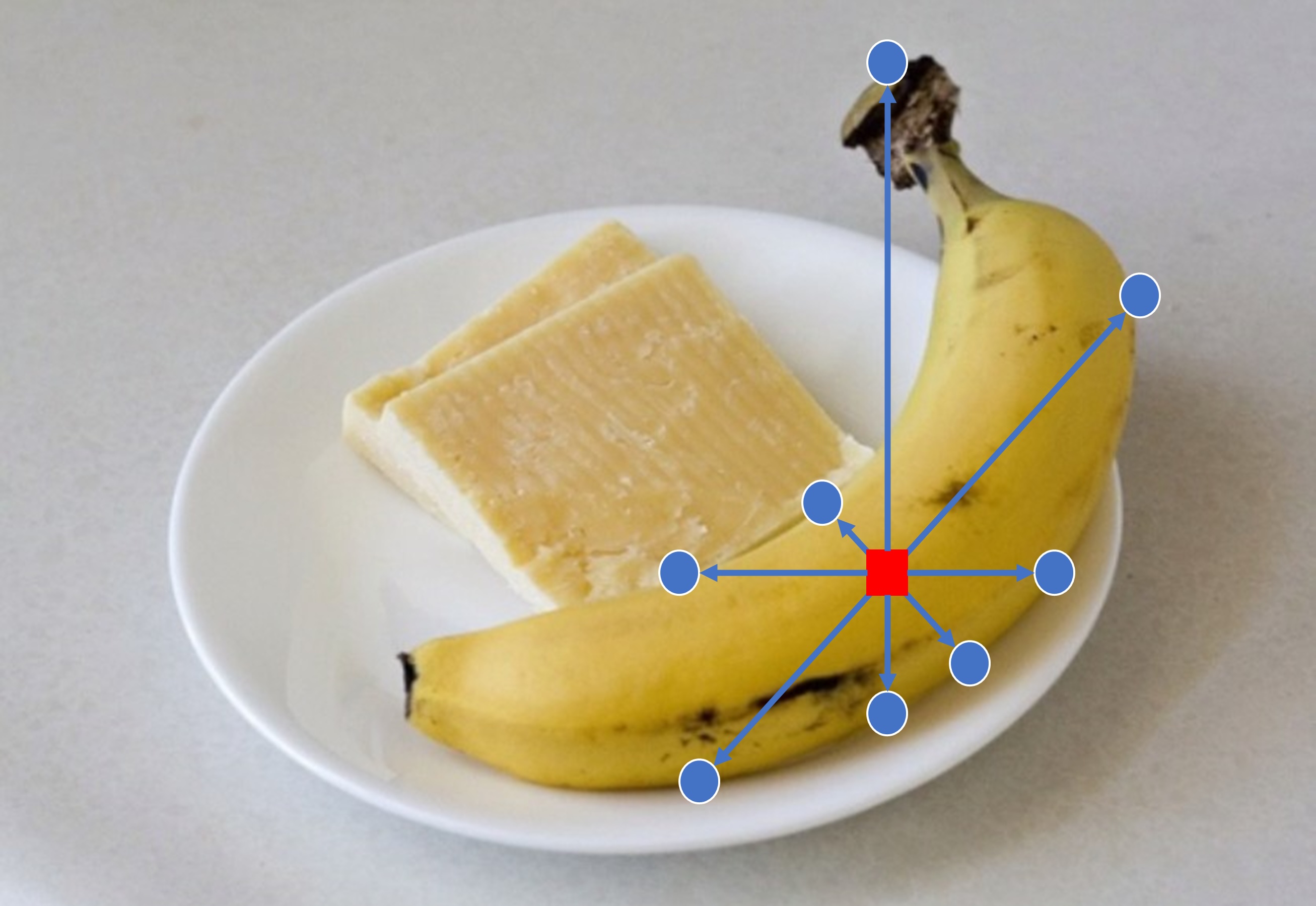}\label{Fig.main2a}}
    \hspace{0.02in}
    \subfloat[Segmentation of PolarMask]
        {\includegraphics[height=1.1in]{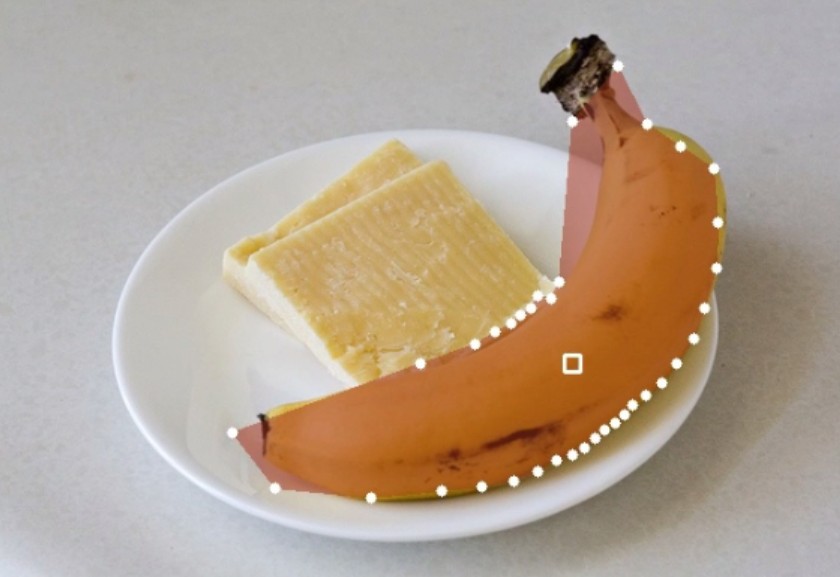}\label{Fig.main2b}}\\
    \vspace{0.1in}
    \subfloat[Mask points of MP-Polarmask]
{\includegraphics[height=1.1in]{figures/banana_a_2.jpg}\label{Fig.main2c}}
    \hspace{0.02in}
    \subfloat[Segmentation of MP-PolarMask]
    {\includegraphics[height=1.1in]{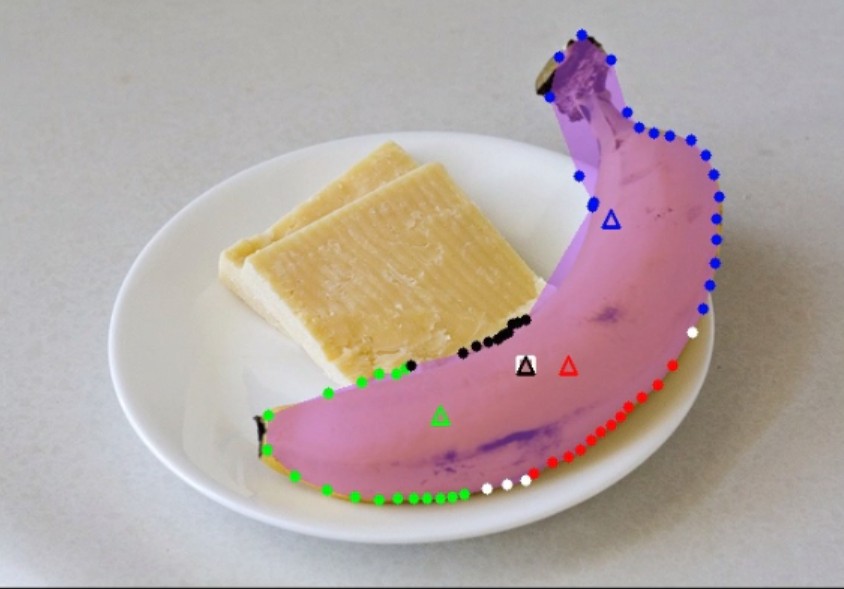}\label{Fig.main2d}}
    \caption{ Comparison of  Polarmask and MP-PolarMask.}
    \label{Fig.main2}
\end{figure}

This paper proposes MP-PolarMask (Multi-Point PolarMask) to relieve the aforementioned problem. We identify a deficiency of the ``Distance Label Generation'' algorithm in PolarMask, which tends to choose longer rays to represent masks, thereby ignoring the internal nodes within an object and adversely affecting its performance on concave objects. We then propose to utilize multiple auxiliary points that form multiple Polar systems to represent a mask. The predicted masks are finer, and there is not much extra computation cost in addition to PolarMask. We have validated MP-PolarMask on multiple datasets, with special focus on the food images in the COCO dataset. \cref{Fig.main2} illustrates the main idea of this work. \cref{Fig.main2a} and \cref{Fig.main2c} show examples of mask points found by PolarMask and MP-PolarMask, respectively, when using 8 rays. \cref{Fig.main2b} and \cref{Fig.main2d} show the exact output masks by PolarMask and MP-PolarMask, respectively, when using 36 rays.

The rest of this paper is organized as follows. Some related works are introduced in \cref{chap2}. \cref{sec3} presents MP-PolarMask. \cref{sec4} shows our experiment results. Conclusions are drawn in \cref{sec5}.

\section{Related Works} \label{chap2}
\subsection{Anchor-based and Anchor-free Detection} \label{sec2.1}

Anchor-based methods are widely used in object detection and instance segmentation tasks. These methods use a predefined set of bounding boxes, known as anchors, to localize and classify objects within an image. The anchor boxes define candidate regions of different shapes and sizes, representing potential object locations and aspect ratios. During training, the model adjusts these anchor boxes to better match the ground-truth bounding boxes of objects in the image and to predict their positions and categories.

Anchor-free methods do not rely on predefined anchor boxes to determine objects' positions. Objects' positions are directly predicted by such networks.
CornerNet \cite{Law2020} and CenterNet \cite{9010985} are two examples of anchor-free methods. CornerNet predicts object positions by estimating the top-left and bottom-right corners of objects. On the contrary, CenterNet takes a different approach by directly forecasting the center points of objects and employing convolutional operations to determine their boundaries.

Anchor-free methods are often simpler than traditional anchor-based methods and can achieve better detection results in certain scenarios. These methods are easier to implement since they eliminate the need for designing and tuning numerous anchor boxes. Despite their advantages, anchor-free methods also face challenges. They might struggle with detecting small and overlapped objects. Additionally, these methods often require more computational resources, creating a hurdle for resource-constrained devices.

\subsection{Instance Segmentation} \label{sec2.2}

Instance segmentation is an important task in computer vision that may foster many downstream tasks. According to whether object detection and segmentation proceed in parallel, instance segmentation can be divided into one-stage and two-stage methods.

Two-stage instance segmentation methods, such as Mask R-CNN \cite{he2018mask}, first generate candidate regions of interests (ROIs) and then classify and segment those ROIs in the second stage. 
Because it requires re-extracting features for each ROI and processing them with subsequent computations, achieving real-time speeds remains challenging.

One-stage instance segmentation methods generate position-sensitive maps that are assembled into final masks by position-sensitive pooling or by combining semantic segmentation and direction prediction logits. In contrast to two-stage approaches, one-stage methods generally offer faster processing speeds at the cost of reduced accuracy. TensorMask \cite{9010024} is a noexception among single-stage methods, as it achieves comparable accuracy to the two-stage Mask R-CNN. 
YOLACT \cite{bolya2019yolact} eliminates the necessity for proposal (bounding-box) generation and feature pooling head networks used in two-stage methods, allowing it to achieve competitive accuracy at real-time speed ($33$ frames per second) on the COCO dataset \cite{10.1007/978-3-319-10602-1_48}. When considering the same image size and device specifications, PolarMask with ResNet-$101$ \cite{7780459} backbone is $4.7$ times faster than TensorMask. The subsequent Polarmask++ \cite{9159936} is proved to be even superior with a significant speed advantage over TensorMask.

\subsection{PolarMask} \label{sec2.3}

PolarMask \cite{9157078} stands out as an unique instance segmentation algorithm that is designed to keep a balance between accuracy and efficiency. Taking a one-stage approach, it performs object detection and segmentation simultaneously. Rather than using a traditional $xy$-coordinate, it is built upon a Polar coordinate system to model a contour. The segmentation problem is thus transformed into two subproblems: {\textit{center regression}} and {\textit{mask ray regression}}.  

\begin{figure} 
\centering 
\includegraphics[width=0.47\textwidth]{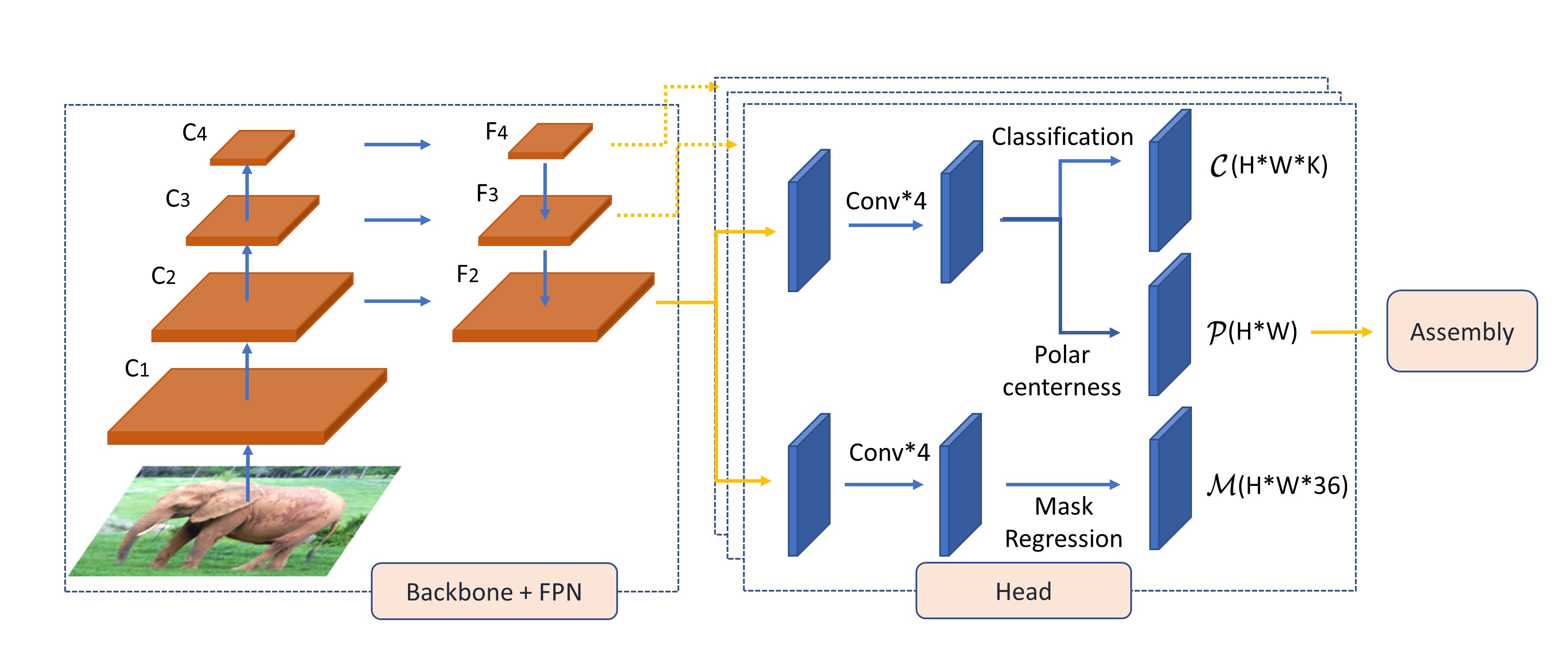} 
\caption{PolarMask \cite{9157078}.} 
\label{Fig.main1} 
\end{figure}

\cref{Fig.main1} shows its architecture. The first part is backbone and FPN. The backbone serves as a feature extractor, which can be realized by different architectures such as ResNet or ResNeXt. FPN then works as a generator to produce multi-scale feature maps through a top-down pathway coupled with lateral connections. We exemplify the idea by three feature maps $F_2, F_3,$ and $ F_4$. From $F_4$, via upsampling, $1 \times 1$ convolutions, and fusing with $C_3$, the feature map $F_3$ is yielded. In a similar way, $F_2$ is yielded. Then a $3 \times 3$ convolution layer is employed to smooth each feature map. The top-down pathway allows variable receptive fields to capture broader and more abstract information.

The second part, Head, has three parallel prediction networks, each for processing one scale of feature map. For each $F_i$, three branches are designed: {\em classification}, {\em Polar centerness}, and {\em mask regression}. $F_i$ first goes through some convolution layers. The classification branch produces a matrix $\mathcal{C} \in R^W \times R^H \times R^k$, where each tensor $(i,j,*)$ is the probability of $k$-class prediction. The Polar centerness branch produces a matrix $\mathcal{P} \in R^W \times R^H$, where each item $(i,j)$ is the score of pixel $(i,j)$ being a Polar center. The mask regression branch computes a matrix $\mathcal{M} \in R^W \times R^H \times R^n$, where each tensor $(i,j,*)$ denotes the lengths of $n$ rays. For example, when $n = 36$, there are $36$ rays, each separated by $360/n = 10$ degrees, whose length is described by $(i,j,k)$ for $k=1, 2, \ldots, 36$. The contour connected by the endpoints of these $36$ rays forms the Polar mask of the object.

\begin{figure} [t]
    \centering
    \begin{subfigure}[b]{\linewidth}
        \centering
    {\includegraphics[height=1in]{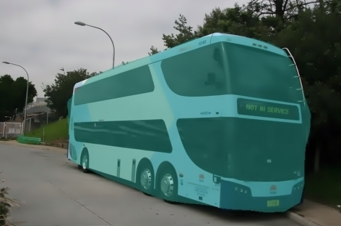}\label{Fig.main3a}}
    \hspace{0.02in}
    {\includegraphics[height=1in]{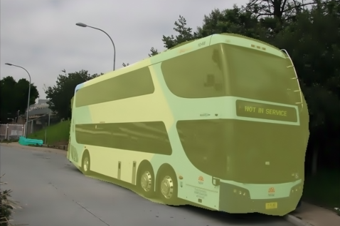}\label{Fig.main3b}}
    \caption{Convex-like object}
    \end{subfigure}
    \begin{subfigure}[b]{\linewidth}
            \vspace{0.1in}
            \centering
{\includegraphics[height=1in]{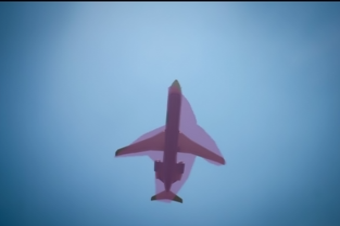}\label{Fig.main3c}}
    \hspace{0.02in}
    {\includegraphics[height=1in]{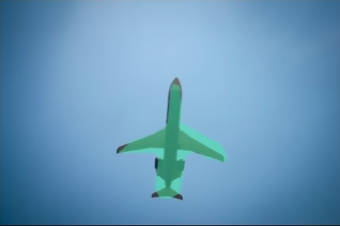}\label{Fig.main3d}}
        \caption{Concave-like object}
    \end{subfigure}
    
    \caption{Left: PolarMask; Right: MP-PolarMask}
    \label{Fig.comparison}
\end{figure}

The third part, Assembly, combines the above predictions by performing a pairwise multiplication $\mathcal{C} \times \mathcal{P}$ to get confidence scores, followed by thresholding for identifying the top-$1000$ centers of each $F_i$. At the end, the top-$1000$ predictions across all three scales are combined, subject to non-maximum suppression (NMS), to get multiple instance segmentation results, each represented by a Polar mask.

We remark that PolarMask follows the design of FCOS \cite{tian2019fcos}, but modifies the prediction networks into Polar representations, i.e., $\mathcal{P}$ and $\mathcal{M}$. It benefits from the Polar representation while keeping computation complexity comparable to FCOS.

\begin{figure*} [tb]
\centering 
\includegraphics[width=0.95\textwidth]{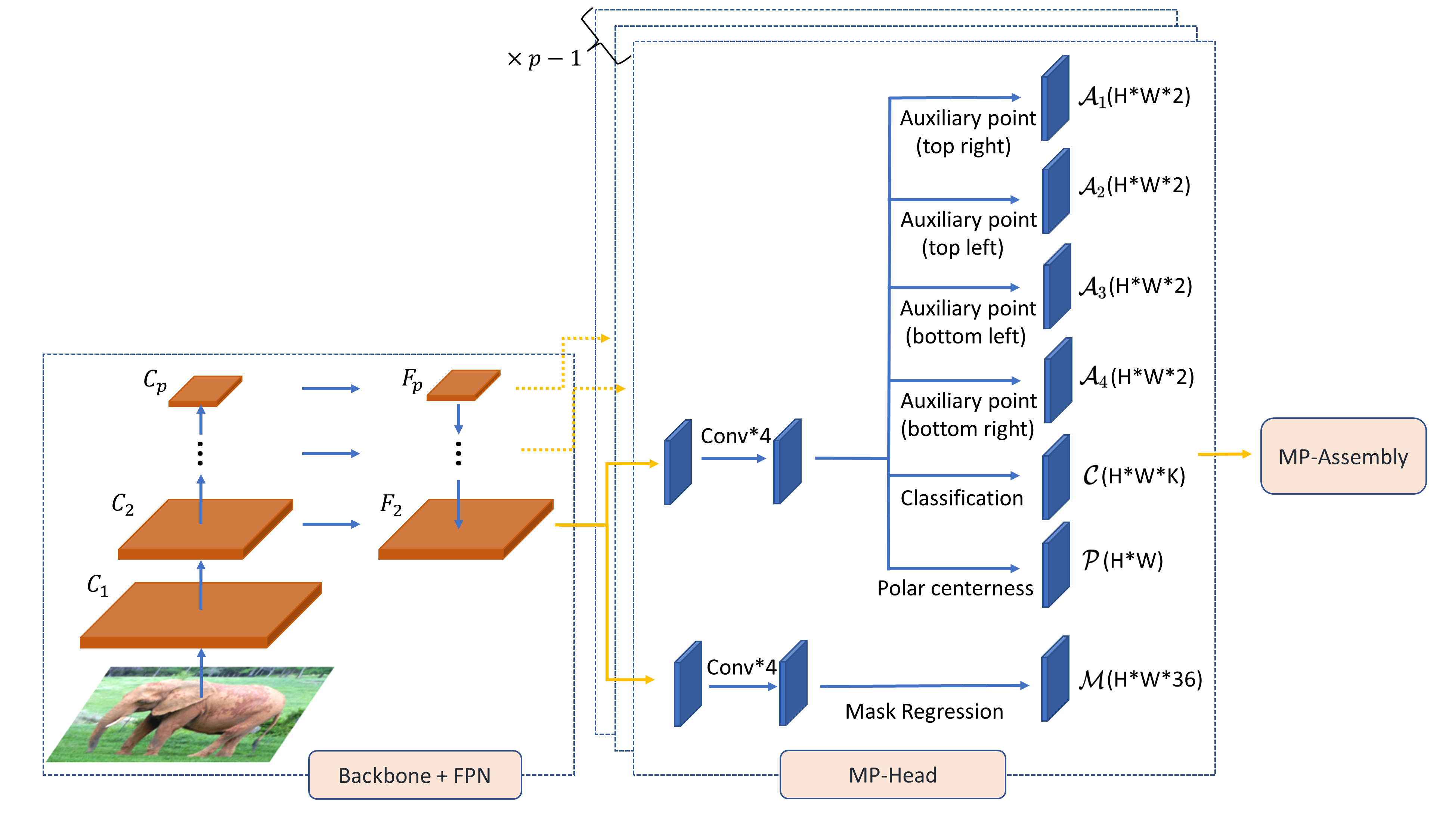} 
\caption{The architecture of MP-PolarMask.} 
\label{Fig.PolarMask} 
\end{figure*}

\section{MP-PolarMask} \label{sec3}

We make two observations on PolarMask. First, using one Polar system, its representation capability is somewhat limited, especially for concave-shaped objects (\cref{Fig.comparison}). In fact, when a ray encounters multiple boundary points of an object, the algorithm would choose the farthest one, tending to form a convex-like polygon. Second, it makes less efficient use of mask regression information. While many Polar centers and rays are identified in $\mathcal{P}$ and $\mathcal{M}$, only one center is selected per object to form the final mask.

To conquer the above deficiencies, we propose Multi-Point PolarMask (MP-PolarMask for short). The main idea is to use multiple Polar systems to represent an object's mask. In \cref{Fig.main2b}, there are actually $5$ Polar systems. A mask is formed by a two-level hierarchy.  First, a main Polar center is determined for an object, from which $n$ rays are defined. Second, from the main center, four quadrants, denoted as $Q_m$, $m=1,2, \ldots, 4$, are defined. Then an auxiliary Polar center is determined for each $Q_m$, from which $n$ additional rays will be extended. These $5n$ rays are assembled to form the final mask.

\begin{table} [htb]

 \centering
\begin{tabular}{c l }
\toprule[2pt]
    {\bf Symbol} & {\bf Description}  \\       \midrule[2pt]
    \multirow{2}{*}{$\mathcal{C}$} &  output matrix of classification branch \\
    & ($R^W \times R^H \times R^k$) \\
    \cmidrule(lr){1-2} 
    \multirow{2}{*}{$\mathcal{P}$} & output matrix of Polar centerness branch  \\ 
    & ($R^W \times R^H$)\\
    \cmidrule(lr){1-2} 
    \multirow{2}{*}{$\mathcal{M}$} & output matrix of mask regression branch  \\
    & ($R^W \times R^H \times R^n$) \\
    \cmidrule(lr){1-2} 
    $F_i$ & feature map, $i = 2, 3, \ldots, p$ \\
    \cmidrule(lr){1-2} 
    \multirow{2}{*}{$\mathcal{A}_m$} & the $m$-th auxiliary-point matrix \\
    & ($R^W \times R^H \times R^2$) \\
    \cmidrule(lr){1-2}
    $\mathcal{Q}_m$ & the $m$-th quadrant\\
    \cmidrule(lr){1-2}  
    $(i_0, j_0)$ & the main center of an object\\
    \cmidrule(lr){1-2}  
    $(i_k, j_k)$ & the $k$-th auxiliary center of a main center\\
    \cmidrule(lr){1-2}
    \multirow{3}{*}{$X_m$} 
    & a sequence of $n$ rays extended from the \\ 
    & main center or an auxiliary center  \\ 
    & ($R^m \times R^n$)\\
    \cmidrule(lr){1-2}
    \multirow{2}{*}{$X'_m$} 
    & a sequence of rays extended from an\\ 
    &  auxiliary point falling in $\mathcal{Q}_m$\\
    \cmidrule(lr){1-2}
    $X_{m,k}$ & the $k$-th mask point of $X_m$\\
    \cmidrule(lr){1-2}
    $A_{m, k}$ & the angle to each point $X_{m, k}$ of $X_m$ \\
    \cmidrule(lr){1-2}
    \multirow{2}{*}{$X_0^{Q_i \rightarrow Q_{j}}$} 
    & the mask points between $X'_i$ and $X'_{j}$ \\
    & contributed by the main center\\
    \cmidrule(lr){1-2}
    $L$ & the loss function of MP-PolarMask\\
    \cmidrule(lr){1-2}
    $L_{cls}$ & the instance center classification loss\\
    \cmidrule(lr){1-2}
    \multirow{2}{*}{$L_{reg}$} 
    & the Polar coordinate distance regression \\ 
    & loss\\
    \cmidrule(lr){1-2}
    $L_{st}$ & the structure centerness loss\\
    \cmidrule(lr){1-2}
    $L_{ac}$ & the auxiliary centerness loss\\
\bottomrule[2pt]
\end{tabular}
\caption{Summary of symbols.}
\label{tab:symbol}
\end{table}

\cref{Fig.PolarMask} shows the architecture of MP-PolarMask. The backbone remains the same as PolarMask, which computes $p-1$ scales of feature maps $F_i, i=2, 3, \ldots, p$. The Multi-Point Head (MP-Head) module is designed to compute the main Polar center and its four auxiliary centers located at its four quadrants. Each auxiliary center is also accompanied by $n$ rays. At the end, the Multi-Point Assembly (MP-Assembly) module integrates them into an instance segmentation. Below, we introduce these modules and the ground truth generation procedure. The symbols used in the paper are listed in \cref{tab:symbol}.

\subsection{MP-Head} \label{sec3.2}

The MP-Head module also has $p-1$ parallel networks, each for processing one feature map $F_i, i=2, 3,  \ldots, p$. Each network has $7$ branches: (i) one classification branch, (ii) one Polar centerness branch, (iii) one mask regression branch, and (iv) four auxiliary-center branches. Similar to PolarMask \cite{9157078}, the first three branches compute the matrices 
$\mathcal{C} \in R^W \times R^H \times R^k$, 
$\mathcal{P} \in R^W \times R^H$, 
and
$\mathcal{M} \in R^W \times R^H \times R^n$, respectively.
The $m$-th auxiliary-center branch computes a matrix $\mathcal{A}_m \in R^W \times R^H \times R^2$, $m =1, 2, \ldots, 4$, in which each tensor $(i, j, *) \in \mathcal{A}_m$ is a 2D displacement vector with respect to $(i, j)$ to define the auxiliary center in $Q_m$. 

Specifically, to get the main center, we multiply $\mathcal{C}$ and $\mathcal{P}$. Following the mechanism of PolarMask, we can get a point, say $(i_0, j_0)$, as the main center. From $(i_0, j_0)$, we derive four auxiliary centers:
\begin{equation}
\! \left\{
\begin{array}{l}
   (i_1, j_1) = (i_0 + \mathcal A_1(i_0,j_0,1),
   j_0 + \mathcal A_1(i_0,j_0,2)) \\
   (i_2, j_2)  = (i_0 - \mathcal A_2(i_0,j_0,1),
   j_0 + \mathcal A_2(i_0,j_0,2)) \\
   (i_3, j_3)  = (i_0 - \mathcal A_3(i_0,j_0,1),
   j_0 - \mathcal A_3(i_0,j_0,2)) \\
   (i_4, j_4)  = (i_0 + \mathcal A_4(i_0,j_0,1),
   j_0 - \mathcal A_4(i_0,j_0,2)) 
\end{array} \right.
\label{eq:StructureNess}
\end{equation}
which are located in $Q_1$, $Q_2$, $Q_3$, and $Q_4$, respectively. Further, we use $\mathcal{M}$, the output of the mask regression branch, to obtain $n$ ray lengths, i.e., $\mathcal{M} (i_m, j_m, *)$, for auxiliary center $(i_m, j_m)$. Including the $n$ rays defined by $\mathcal{M} (i_0, j_0, *)$ for the main center $(i_0, j_0)$, we have totally $5n$ rays.

\subsection{MP-Assembly} \label{sec3.3}

This module aims to construct the final mask. The algorithm is outlined in  \cref{alg:MAM}. The inputs include: (i) the main center $(i_0,j_0)$, (ii) four auxiliary centers $(i_m,j_m), i = 1,2, \ldots,4$, and (iii) the tensor $\mathcal{M}$ that defines ray lengths for these centers. The output is a sequence of points that defines the mask of the object.

\begin{algorithm} 
\caption{MP-Assembly}
\label{alg:MAM}
\SetKwInOut{Input}{input}
\SetKwInOut{Output}{output}

\Input{Center $(i, j)$; Auxiliary centers $(i_m, j_m)$, $m=1, 2,\ldots,4$; Tensor $\mathcal{M}$; Number of rays $n$;}
\Output{A sequence of mask points;}

\BlankLine
\SetKwProg{Fn}{Function}{}{end}
\Fn{MP-Assembly()}{
Calculate sequences $X_m$ for $m=0,1, \ldots,4$;
\\
Calculate angle sequences $A_m$ of $X_m$ for $m=1,2, \ldots,4$;
\\
Calculate angles $\alpha_m$ for $m=1,2, \ldots,4$;
\\
Refine $X_m$ into $X'_m$, $m=1,2, \ldots,4$, by removing out-of-angle points;
\\
Refine $A_m$ into $A'_m$, $m=1,2, \ldots, 4$, accordingly;

\For{$m = 1:4$}{
    $a_m \leftarrow \min \{ A'_m \}$;
    \\
    $b_m \leftarrow \max \{ A'_m \}$;
    
}

Calculate the sub-sequences 
$X_0^{Q_1 \rightarrow Q_2}$,
$X_0^{Q_2 \rightarrow Q_3}$,
$X_0^{Q_3 \rightarrow Q_4}$, and 
$X_0^{Q_4 \rightarrow Q_1}$
from $X_0$;
\\
\Return
$(X'_1 | 
X_0^{Q_1 \rightarrow Q_2} |
X'_2 |
X_0^{Q_2 \rightarrow Q_3} |
X'_3 |$
\\
$X_0^{Q_3 \rightarrow Q_4} |
X'_4 |
X_0^{Q_4 \rightarrow Q_1})$;
}
\end{algorithm}

First, we will compute the mask points specified by the main center and the four auxiliary centers, denoted by $5$ sequences $X_m, m=0,1, \ldots,4$, respectively (line 2). The $k$th mask points of these sequences, $k=1, 2, \ldots, n$,  are defined as
\begin{equation}
X_{m, k}  = (i_m, j_m) + \Vec{u}_k \cdot \mathcal{M} (i_m, j_m, k)
\end{equation}
where $\Vec{u}_k$ is a unit vector with direction $2 \pi \cdot (k-1)/ n$. That is, $X_m = \{X_{m, k} \mid {k=1,2, \dots, n}\}$, $m=0,1, \ldots,4$. 

We are going to form a mask from the points in $X_m, m=0,1, \ldots,4$. To do so, we associate an angle to each point $X_{m, k}$ of $X_m$ with respect to the main center:
\begin{equation}
A_{m, k} = \angle X_{m, k} O O^+
\end{equation}
where $O=(i_0, j_0)$ is the main center and $O^+ = (i_0+1, j_0)$ is to form the positive x-axis from the main center. We denote by $A_m$ the angle sequence of $X_m, m=0,1, \ldots,4$ (line 3).

Next, we need to refine the four sequences formed by the auxiliary centers. We identify $4$ points in $X_0$ that divide the $4$ quadrants:
$$
X_{0, 1}, X_{0, 1+n/4}, X_{0, 1+2n/4}, X_{0, 1+3n/4}
$$
For $k = 1, 2, \ldots, 4$, we draw $4$ angles as follows (line 4):
\begin{equation}
\alpha_k  = \angle X_{0, 1 + (k-1) \times n/4} (i_k, j_k) X_{0, 1 + (k \! \! \! \mod 4) \times n/4}
\end{equation}
We take the sub-sequence of $X_m, m=1,2, \ldots, 4,$ that falls within the angle $\alpha_m$ with respect to $(i_m, j_m)$ (line 5). That is
\begin{equation}
X'_m  = subseq(X_m, \alpha_m).
\end{equation}
where function $subseq()$ is to retrieve a sub-sequence within an angle. We also refine $A_m$ into $A'_m$ accordingly (line 6).

\begin{figure*}
\centering
    \subfloat[]
    {\includegraphics[height = 1.8in]
    {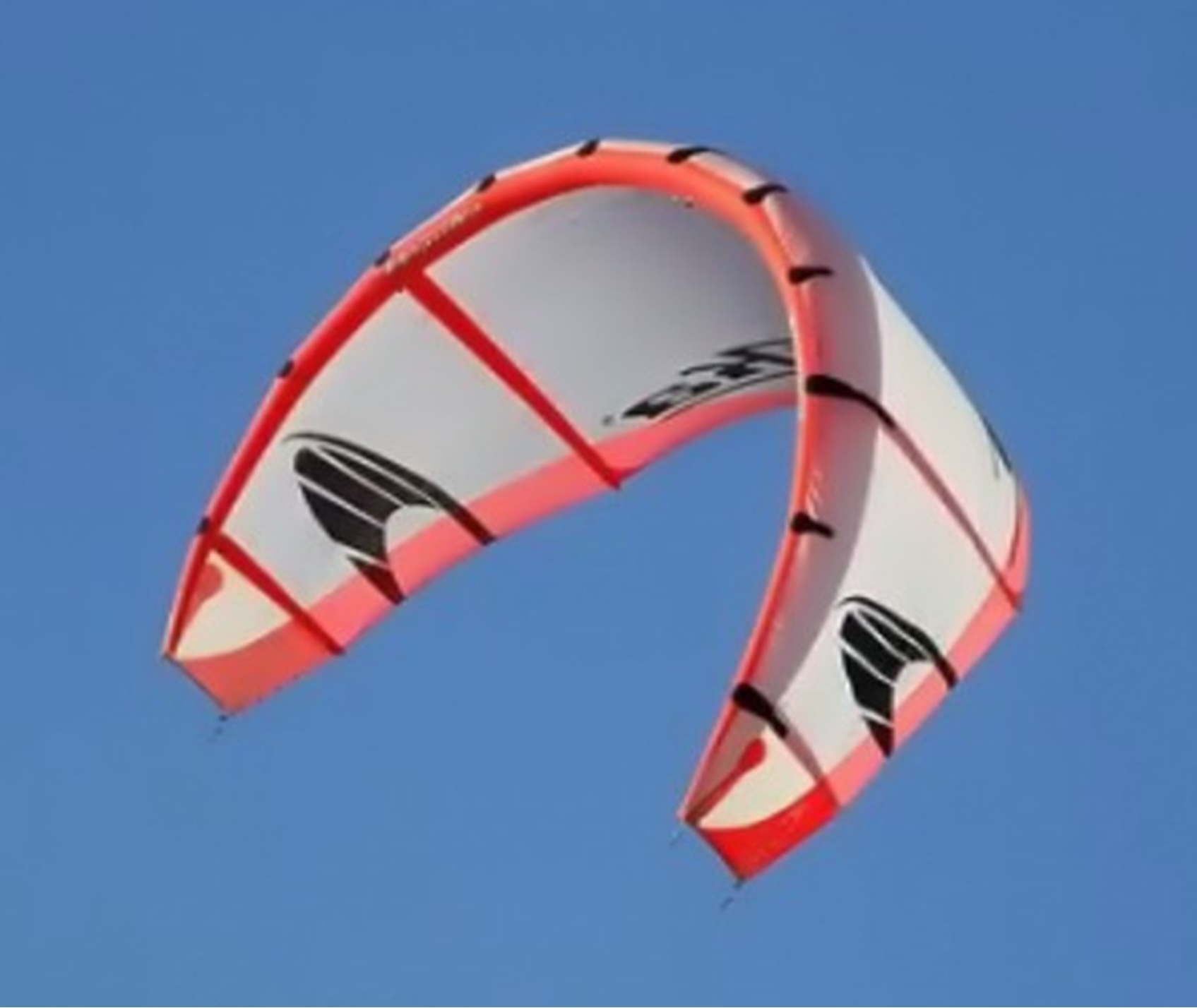}\label{Fig.main4a}}
    \hspace{0.1in}
    \subfloat[]
    {\includegraphics[height = 1.8in]
    {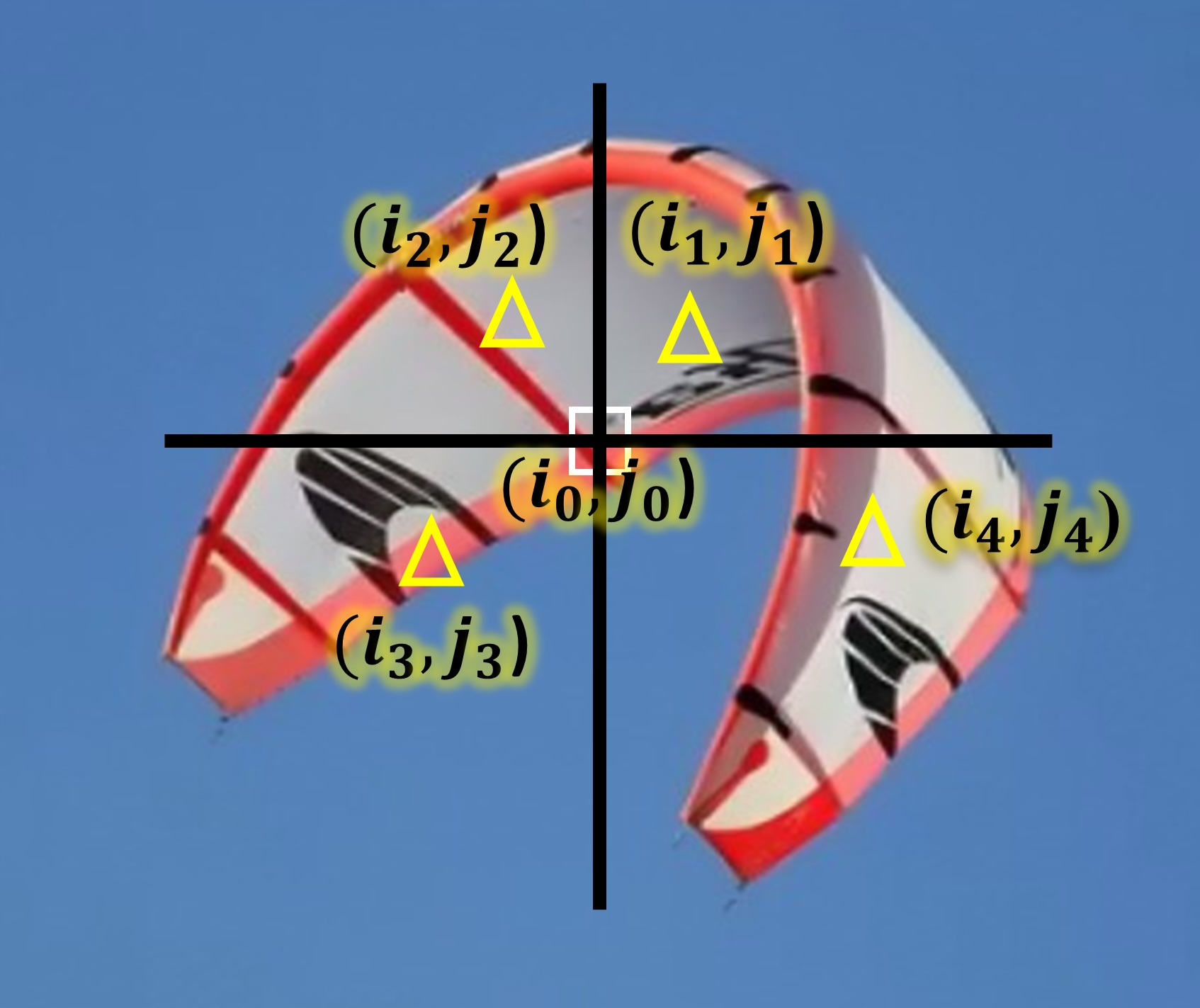}\label{Fig.main4c}}
    \hspace{0.1in}
    \subfloat[]
    {\includegraphics[height = 1.8in]
    {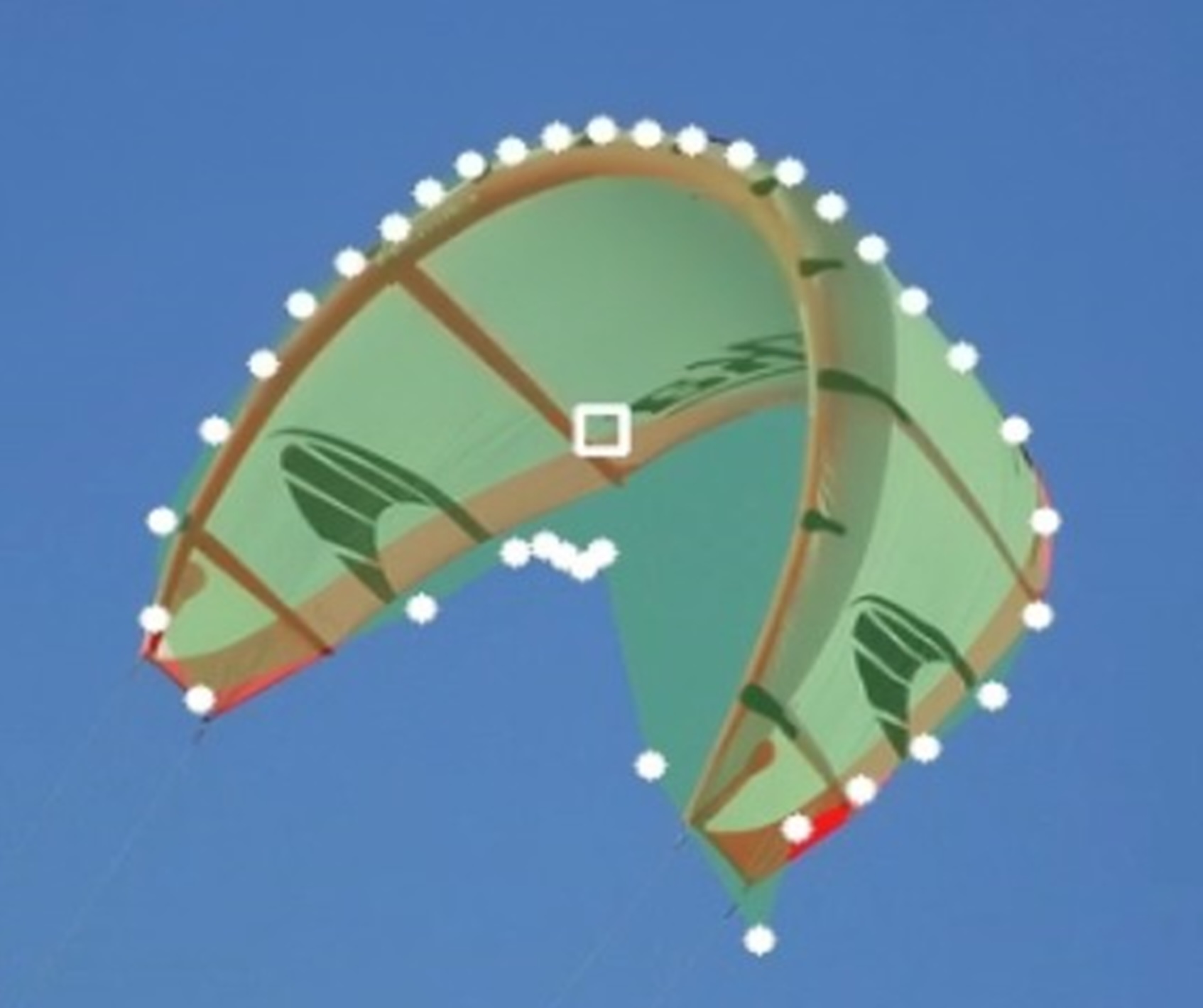}\label{Fig.main4b}}\\
    \vspace{0.1in}
    
    \subfloat[]
    {\includegraphics[height = 1.8in]
    {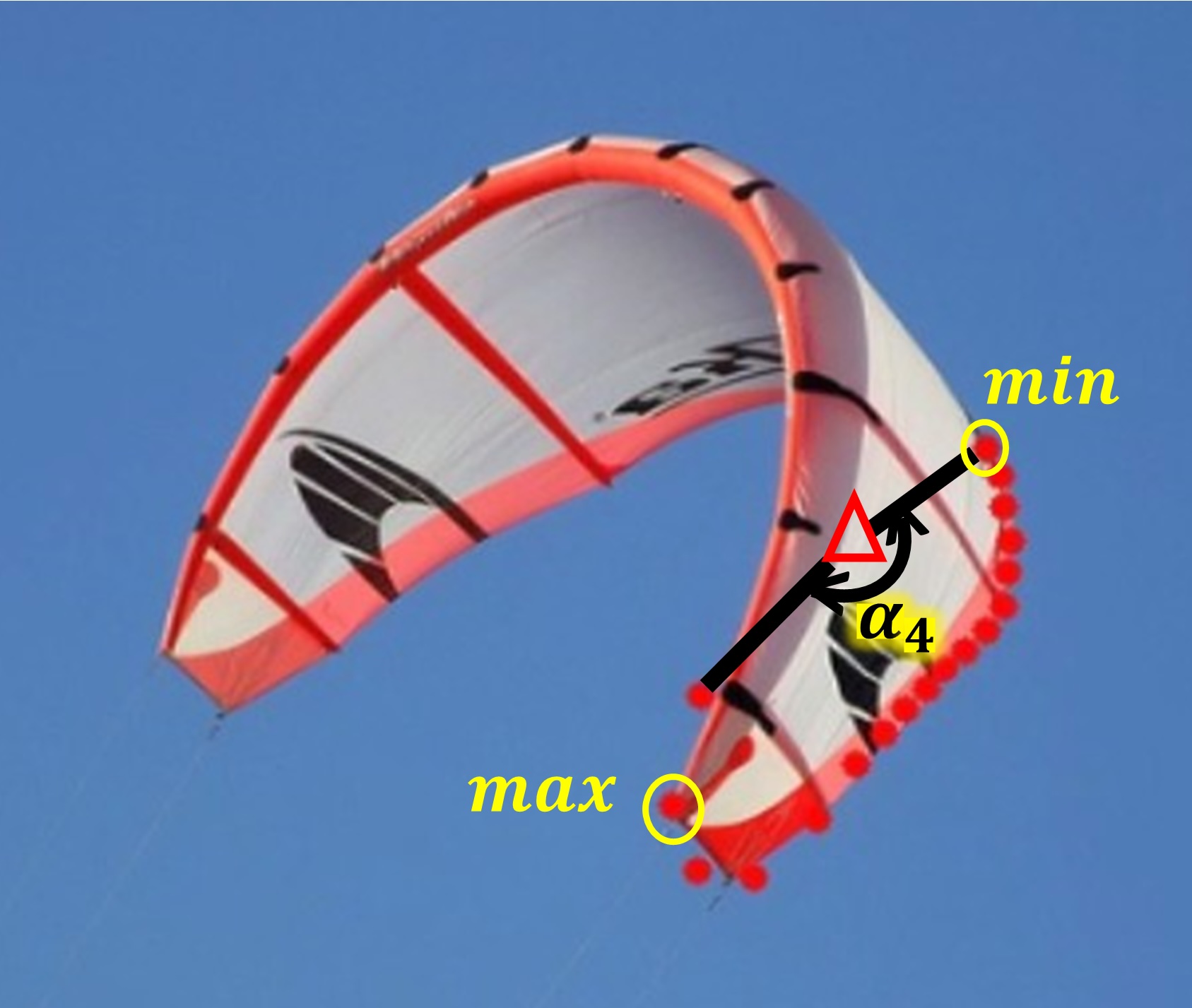}\label{Fig.main4d}}
    \hspace{0.1in}
    \subfloat[]
    {\includegraphics[height = 1.8in]
    {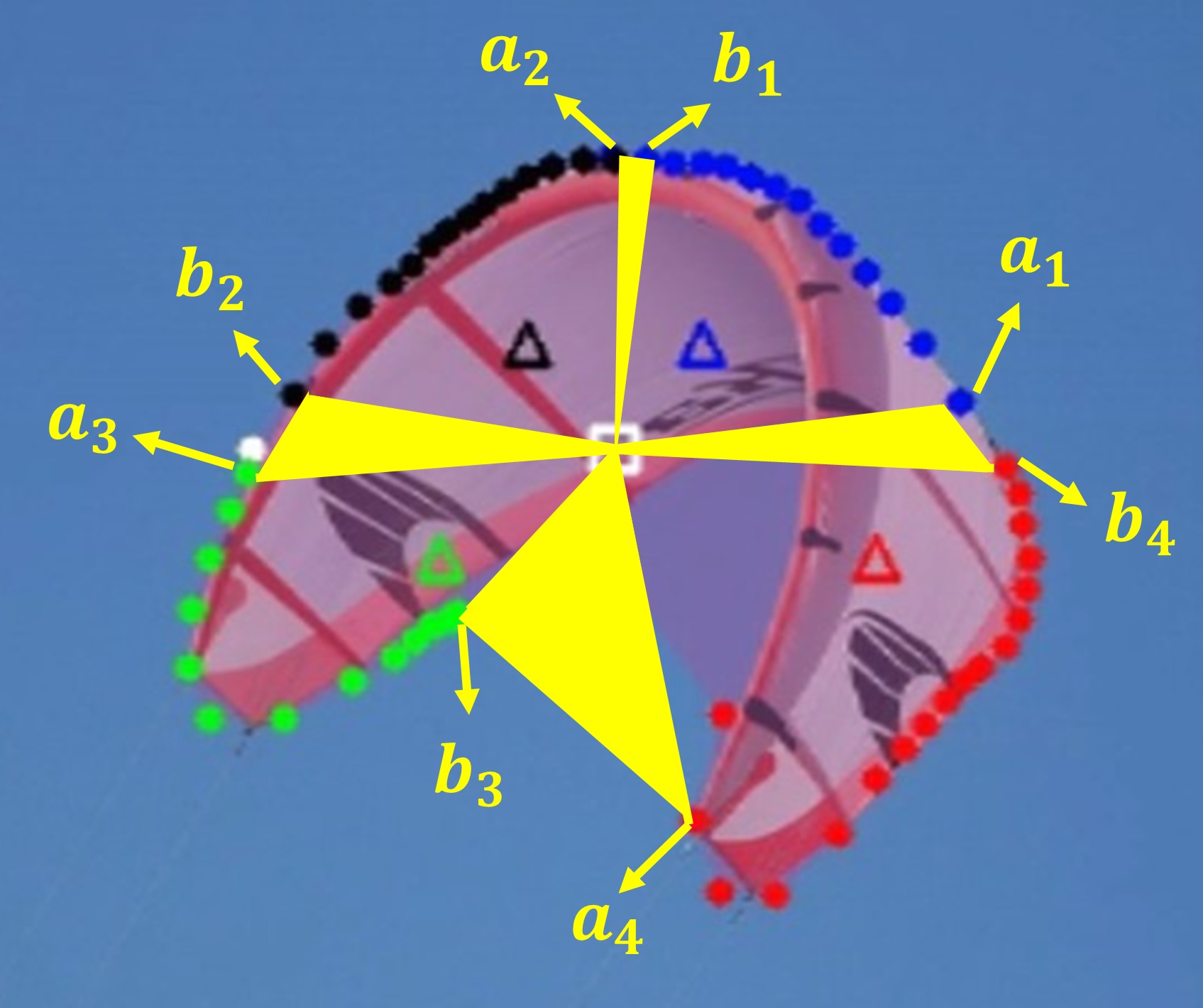}\label{Fig.main4e}}
    \hspace{0.1in}
    \subfloat[]
    {\includegraphics[height = 1.8in]
    {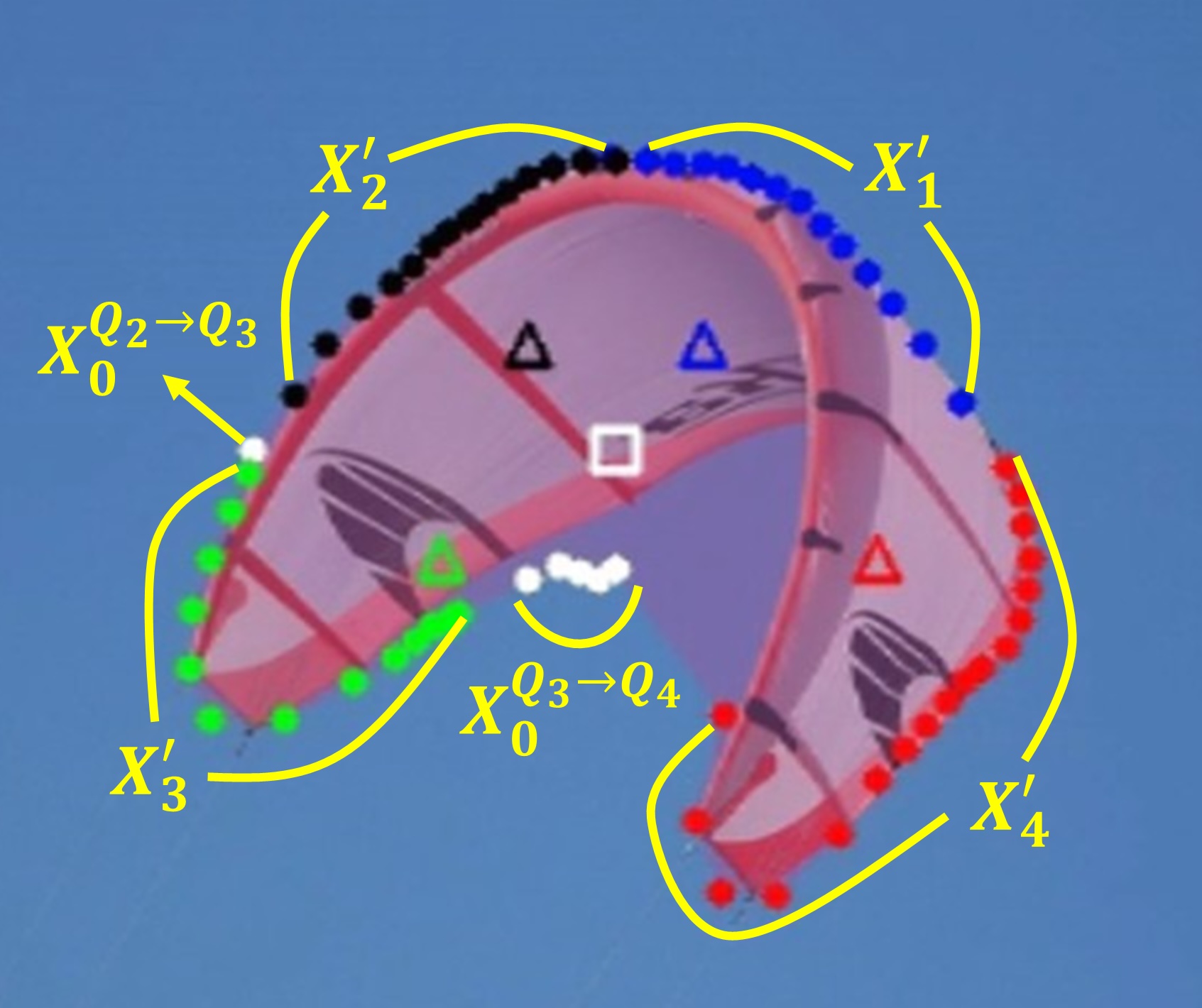}\label{Fig.main4f}}
    \caption{A running example of MP-PolarMask: (a) the input image, (b) the main center and four auxiliary centers,  (c) the mask points expanded from the main center, (d) the sequence $X'_4$ in Quadrant $4$ refined by the angle $\alpha_4$, (e) ($a_m, b_m$) and its corresponding $X'_m$, where the yellow regions are to be filled by the mask points of the main center, and (f) the final mask  $X'_1 | X'_2 | X_0^{Q_2 \rightarrow Q_3} | X'_3 | X_0^{Q_3 \rightarrow Q_4} | X'_4$.}
    \label{Fig.main4}
\end{figure*}

From $A'_m, m=1, 2, \ldots, 4$, we identify the minimal and the maximal angles in the sequence (note that the angles are relative to the main center). These two points are denoted as $a_m$ and $b_m$ (line 7).

In order to integrate the points of $X_0$ with those in $X'_m, m=1,2, \ldots,4$, the final step is to identify the gap between $X'_1$ and $X'_2$, the gap between $X'_2$ and $X'_3$, etc. We derive
\begin{equation}
\left\{
\begin{array}{l}
X_0^{Q_1 \rightarrow Q_2} = subseq(X_0, \angle b_1 O a_2)
\\
X_0^{Q_2 \rightarrow Q_3}  = subseq(X_0, \angle b_2 O a_3)
\\
X_0^{Q_3 \rightarrow Q_4}  = subseq(X_0, \angle b_3 O a_4)
\\
X_0^{Q_4 \rightarrow Q_1}  = subseq(X_0, \angle b_4 O a_1)
\end{array} \right.
\end{equation}
where $O$ is the main center (line 11). The final mask is formed by concatenating $X'_m, m=1,2, \ldots,4$, and the above four sub-sequences (line 12). 

Below, we use \cref{Fig.main4} to run an example. \cref{Fig.main4a} shows an image with an concave-shaped object. By running MP-PolarMask, the main center and the four auxiliary centers in four Quadrants are identified in \cref{Fig.main4c}. These centers further identify five sequences $X_m, m=0..4$. In \cref{Fig.main4b}, the white points are the potential mask points expanding from the main center, i.e., $X_0$. In fact, these points would form the segmentation result found by PolarMask. We can observe that PolarMask misses a lot of areas in Quadrant 4. \cref{Fig.main4d} shows the angle $\alpha_4$ that is determined by the maximum and the minimum angles of the points in $X_4$ in Quadrant 4. The points falling within $\alpha_4$ constitute the sequence $X'_4$. \cref{Fig.main4e} demonstrates the angle pairs ($a_m, b_m$) in all Quadrants, delineating the sequences $X'_m$ and the gaps between them. The final mask prediction is combined by $X'_m$ and $X_0^{Q_i \rightarrow Q_j}$ as shown in \cref{Fig.main4f}.

\subsection{Loss Functions} \label{sec3.4}

We formulate the loss function as follows
\begin{equation}
    L = L_{cls} + L_{reg} + L_{sc} + L_{ac}
    \label{eq:Loss}
\end{equation}
where $L_{cls}$ is the instance center classification loss,
$L_{reg}$ is the Polar coordinate distance regression loss, 
$L_{sc}$ is the structure centerness loss, and 
$L_{ac}$ is the auxiliary centerness loss. We extend the losses in PolarMask \cite{9157078} to cover auxiliary centers. $L_{cls}$ is formulated as the focal loss function \cite{lin2018focal}, so we omit the details. We explain the other three terms below.

Following PolarMask, MP-PolarMask transforms the task of instance segmentation into a set of regression problems. In most object detection and image segmentation tasks, smooth L1 loss and IoU loss are two effective methods to supervise regression tasks. Smooth L1 loss ignores the correlation between samples of the same object, which leads to lower localization accuracy, while IoU loss considers the optimization globally and directly optimizes the pixel outcomes. However, computing the IoU of two areas is challenging and hard to parallelize. In Polarmask, it simplifies the computation of IoU by the following distance regression loss:
\begin{equation}
    L_{reg} =  \sum_{(x, y) \in OC} f(x,y)
      \label{eq:polarIoUloss}
\end{equation}
\begin{equation}
    f(x,y) = 
          \log \frac
          {\sum_{i=1}^{n} 
          \max\{ \mathcal{M}(x, y, i), \mathcal{M}^*(x, y, i)\}
          }
          {\sum_{i=1}^{n}
          \min\{ \mathcal{M}(x, y, i), \mathcal{M}^*(x, y, i)\}}
    \end{equation}
where  $OC$ means the set of points belonging to any object class and $\mathcal{M}^*$ means the ground truth ray lengths. This is proved to be quite effective in \cite{9157078}, so we follow the same design.

\begin{figure} 
    \centering 
    \subfloat[]
    {\includegraphics[height = 1.2in]{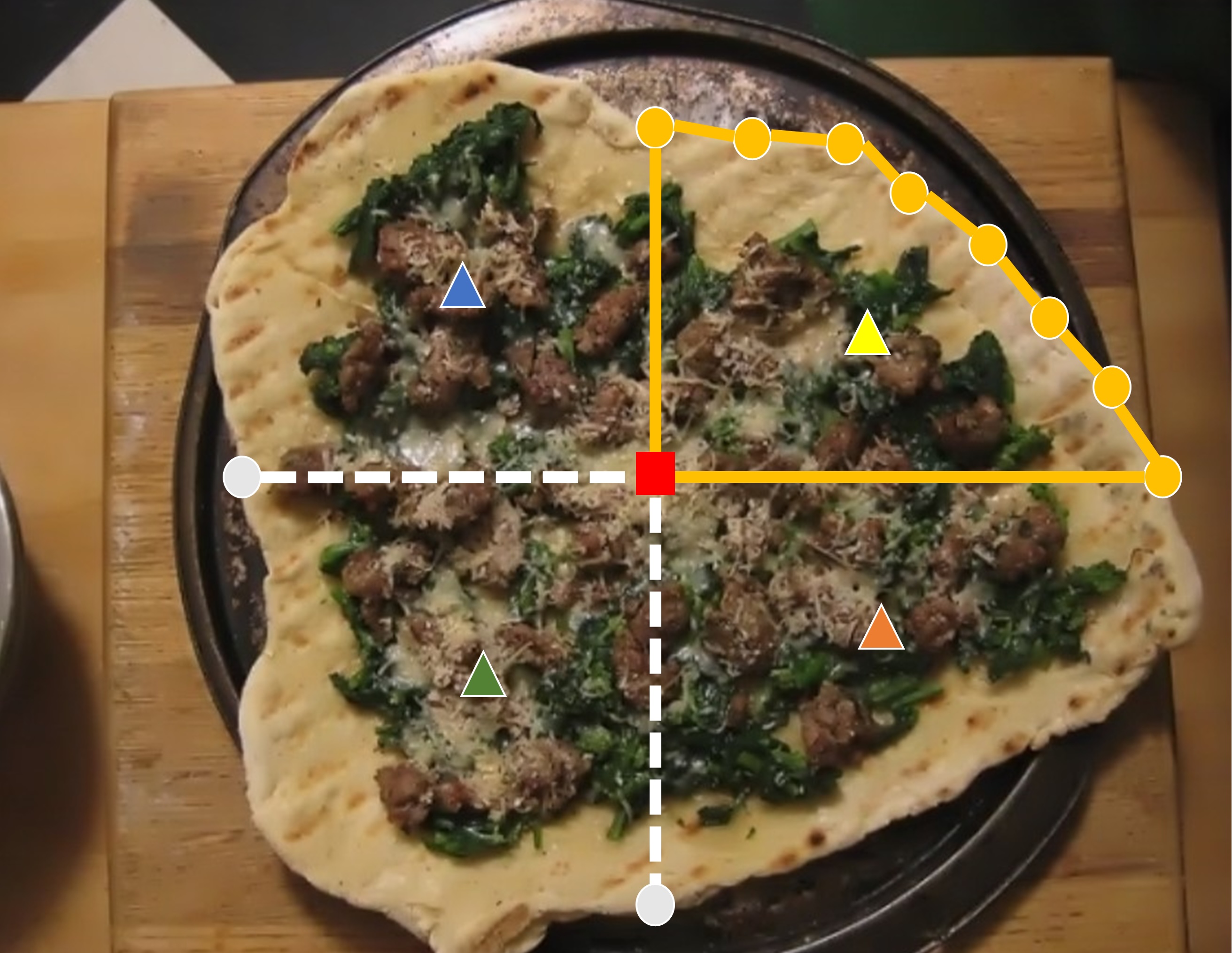}}
    \hspace{0.1in}
    \subfloat[]
    {\includegraphics[height = 1.2in]{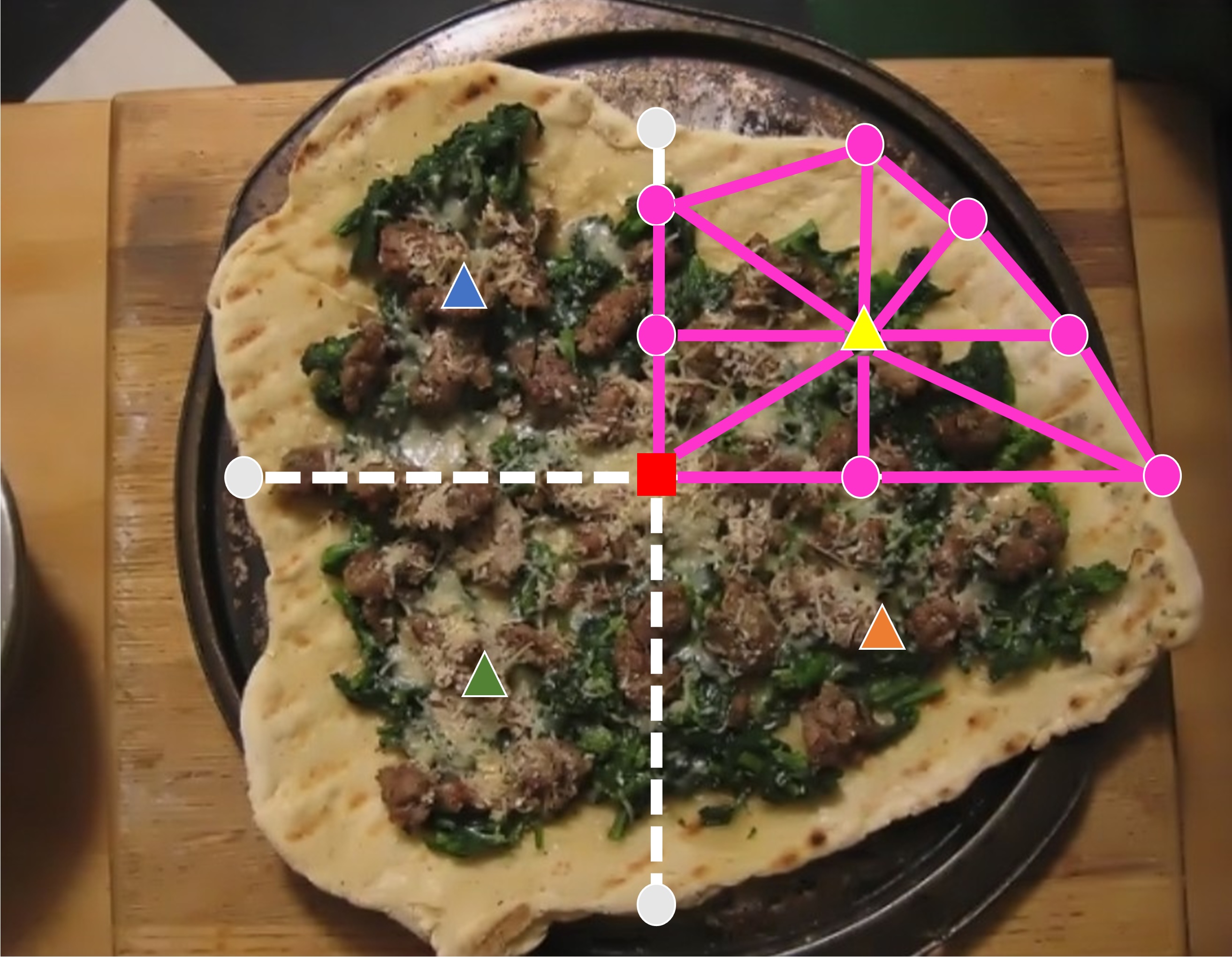}}
    \caption{(a) area $C_1^*(i_0, j_0)$ in Quadrant $1$ for center $(i_0, j_0)$ enclosed by ground truth points and (b) area $C_1(x_1, y_1)$ for Quadrant $1$ for the mass center $(x_1, y_1)$.}
\label{Fig.main6} 
\end{figure}

Polarmask introduces the concept of Polar centerness into its loss     
    \begin{equation}
    \text{PolarCent} = \sqrt{ 
    \frac{\min\{ \mathcal{M}^*(x, y, i) \mid i=1, 2, \ldots, n \}}
    {\max \{ \mathcal{M}^*(x, y, i) \mid i=1, 2, \ldots, n \}}}
    \label{eq:important}
    \end{equation}
where $(x, y)$ is a candidate center point. In our method, since there is a main center and four auxiliary centers, rather than considering centerness as a point, we consider centerness as a ``structure.'' Therefore, we propose the Polar structure centerness as follows. During training, we will compute the matrix $\mathcal{P}$. We will use the ground truth to compute an optimal matrix $\mathcal{P}^*$ and train our model to approximate its output $\mathcal{P}$ to the optimal $\mathcal{P}^*$. The optimal matrix $\mathcal{P}^*$ is computed as follows. Consider any point $(i_0, j_0)$ that belongs to any object class. We use it to partition the object into 4 Quadrants. Let the mask of the object that falls in Quadrant $m$ be $C_m^*(i_0, j_0)$. From the mask, we compute the mass center in Quadrant $m$, denoted by $(x_m, y_m)$. From $(x_m, y_m)$, we further compute a mask that may reflect the best inference result, called $C_m(x_m, y_m)$. Specifically, we take the $n$ rays in the ground truth $\mathcal{M}^*(x_m, y_m, *)$. However, if a ray crosses the $x$-axis or the $y$-axis, it will end at that intersection point; otherwise, the ray remains unchanged. Then, the contour formed by the endpoints of these $n$ rays is $C_m(x_m, y_m)$. In \cref{Fig.main6}, we illustrate the concept using a simple $n=8$ case in Quadrant $1$. So, we define the structure centerness at $(i_0, j_0)$ in an IoU style:
\begin{equation}
\mathcal{P}^* (i_0, j_0)  = \frac{1}{4}
\sum_{i=1}^{4}
\frac{\lvert C_m(x_m, y_m) \cap C^*_m (i_0, j_0) \rvert}
{\lvert C_m(x_m, y_m) \cup C^*_m (i_0, j_0) \rvert}
\label{eq:PolarStructureNess}
\end{equation}
We can repeat the above process for all $(i_0, j_0)$ to obtain the optimal matrix $\mathcal{P}^*$. During training, we calculate the cross entropy loss of the predicted $\mathcal{P}$ and the optimal $\mathcal{P}^*$, denoted as $L_{sc}$.

For the auxiliary center loss, we define the optimal auxiliary array $\mathcal{A}^*_m \in R^W \times R^H \times 2, m=1, 2, \ldots, 4$, as follows:
\begin{equation}
\mathcal{A}^*_m [x, y] = mass_m(x,y) - (x, y)
\end{equation}
where $mass_m(x,y)$ returns the mass center of the object mask in Quadrant $m$ with respect to the origin $(x, y)$. The loss, \cref{eq:Auxiliary loss}, is defined based on the distance between the ground truth points, \cref{eq:dist_loss} and the predicted locations, \cref{eq:aux_arr}.
 \begin{equation}  
        L_{ac} = \sum_{(x, y) \in OC} L_{ac} (x, y)
        \label{eq:Auxiliary loss}
    \end{equation}
\begin{equation}
        dist_m(x, y)  = 
        |\mathcal{A}^*_m[x, y] - 
            \mathcal{A}_m[x, y]|
            \label{eq:dist_loss}
    \end{equation}
 \begin{equation}           
    \!\!\!  L_{ac} (x, y) 
        \! = \! \left\{
            \begin{array}{ll}
            { \! \! 0.5( dist_m(x, y) )^2} & {\! \! \textrm{if }} 
            dist_m(x, y) \! < \! 1 \\
             \! \! dist_m(x, y) - 0.5  &   \! \! \textrm{otherwise}
            \end{array}
            \right .
            \label{eq:aux_arr}
    \end{equation}

\begin{table*} [ht]
  \centering
  \begin{tabular}{@{}lccccccc@{}}
    \toprule[2pt]
    \; &{\bf AP} & {\bf AP$_{50}$}& {\bf AP$_{75}$}& {\bf AP$_{S}$}& {\bf AP$_{M}$}& {\bf AP$_{L}$} & {\bf FPS} \\
    \midrule[2pt]

 \multicolumn{8}{@{}l}{\textbf{Dataset A: All images of COCO test-dev}} \\
 \;\;\;\;Mask R-CNN \cite{8237584} &$35.7$& $58.0$ &$37.8$ &$15.5$& $38.1$ &$52.4$&$7.2$\\
 \;\;\;\;FCIS \cite{8099955} & $29.5$ &$51.5$& $30.2$& $8.0$ &$31.0$& $49.7$&$0.8$\\
 \;\;\;\;YOLACT \cite{bolya2019yolact} &$29.8$& $48.5$ &$31.2$& $9.9$ &$31.3$& $47.7$&$22.1$\\
 \;\;\;\;Tensormask \cite{9010024} & \underline{37.1} & \underline{59.3} & \underline{39.4} & \underline{17.1} &$39.1$ &$51.6$ &$2.1$\\
\;\;\;\;Extremenet \cite{8954244} & $18.9$ & $44.5$ & $13.7$ & $10.4$ &$20.4$& $28.3$&$2.2$\\
 \;\;\;\;PolarMask(36 rays) \cite{9157078} & $32.1$ &$53.7$ &$33.1$& $14.7$ &$33.8$ &$45.3$&$13.9$\\
 \;\;\;\;PolarMask(side $=600$, 36 rays) \cite{9157078} & 30.7 &52.1 &31.9& $13.2$ &$31.9$ &$43.4$& {\bf 23.2} \\
\;\;\;\;MP-PolarMask(36 rays, ResNet-101) & $35.5$ & $58.5$&  $36.6$&$16.2$ &$37.7$ & $51.7$&$13.3$\\
 \;\;\;\;MP-PolarMask(side $=600$, 36 rays, ResNet-101) & $34.8$ & $57.8$&  $35.9$& $15.4$ & $37.2$ & $51.1$& \underline{22.8} \\
\;\;\;\;MP-PolarMask(36 rays, BFP) & $35.6$ & $58.8$&  $36.8$&$16.5$ & \underline{39.4} & \underline{52.8} &$13.2$\\
 \;\;\;\;MP-PolarMask(36 rays, DCN) & {\bf 37.5} & {\bf 60.3} & {\bf 39.6} & {\bf 17.5} & {\bf 39.7} & {\bf 53.5} &$9.2$\\
\cmidrule(lr){1-8}
 \multicolumn{8}{@{}l}{\textbf{Dataset B: Food images of COCO test-dev}} \\

 \;\;\;\;PolarMask($36$ rays) \cite{9157078} & $31.8$ &$51.2$ &$32.0$& $13.1$ &$32.6$ &$43.8$ & {\bf 14.0} \\
 \;\;\;\;MP-PolarMask(36 rays, ResNet-101) & \underline{34.1} & $56.4$& $33.2$&$14.8$ &$35.7$ & $49.8$ & \underline{13.4} \\
 \;\;\;\;MP-PolarMask(36 rays, BFP) & \underline{34.1} & \underline{56.6} & \underline{33.3} & \underline{14.9} & \underline{36.0} & \underline{50.1} &$13.3$\\
 \;\;\;\;MP-PolarMask(36 rays, DCN) & {\bf 35.8} & {\bf 59.3} & {\bf 36.2} & {\bf 16.2} & {\bf 37.3} & {\bf 51.2} &$9.2$\\

    \bottomrule[2pt]
  \end{tabular}
\caption{Performance comparison of MP-PolarMask and other methods. (best boldfaced and second best underlined)} 
\label{tab2}
\end{table*}

    \begin{figure*}
    \centering 
    {\includegraphics[height=0.62in]
    {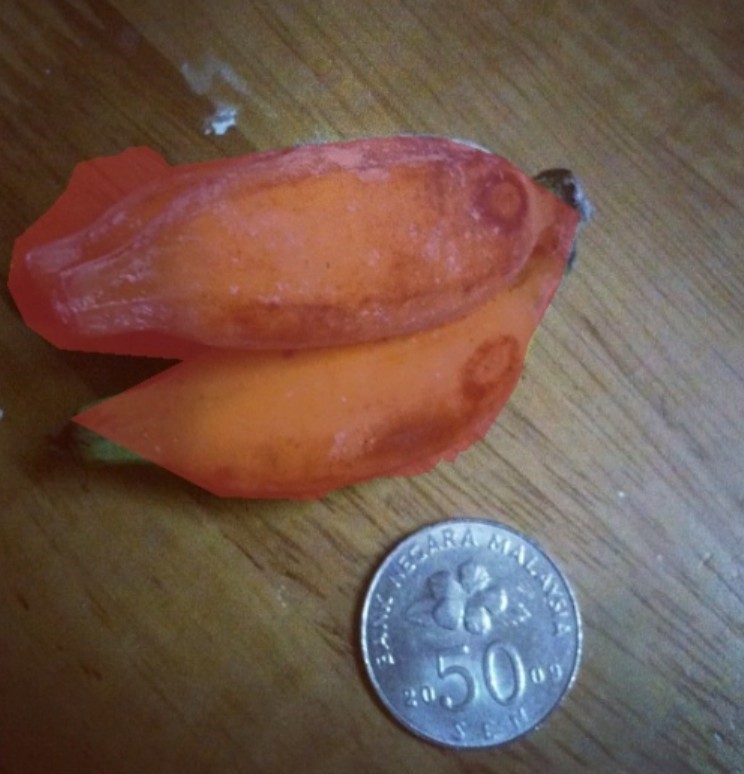}}
    {\includegraphics[height=0.62in]
    {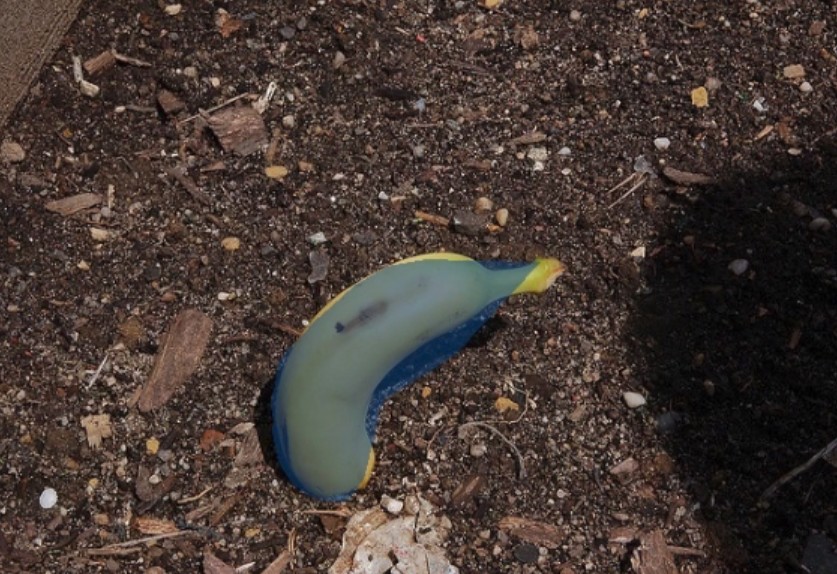}}
    {\includegraphics[height=0.62in]
    {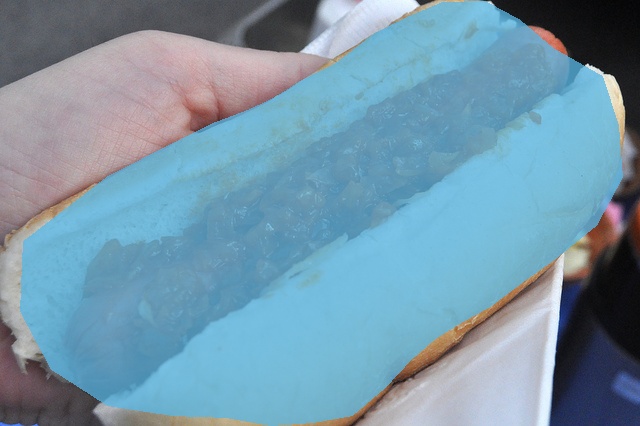}}
    {\includegraphics[height=0.62in]
    {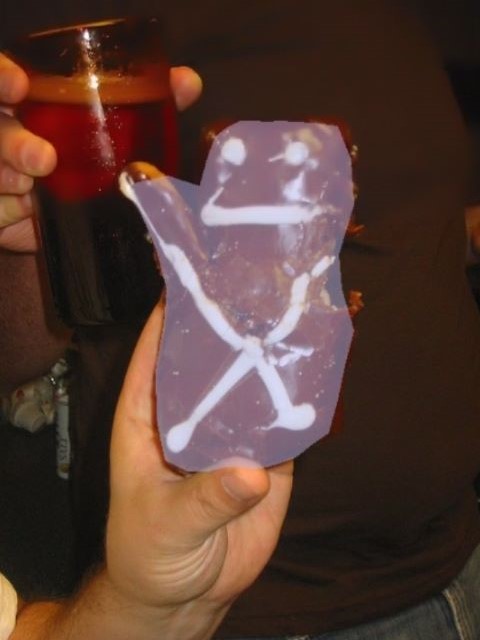}}
    {\includegraphics[height=0.62in]
    {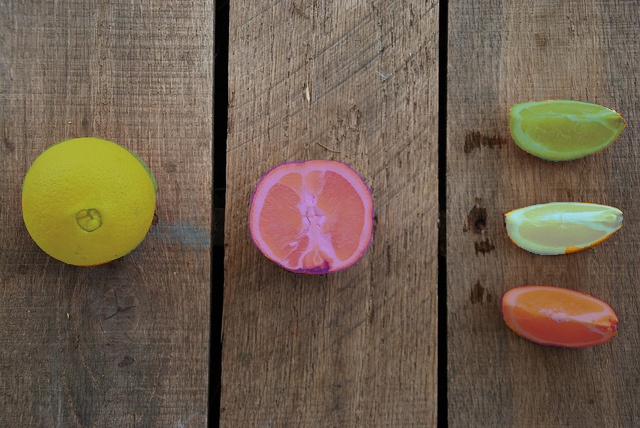}}
    {\includegraphics[height=0.62in]
    {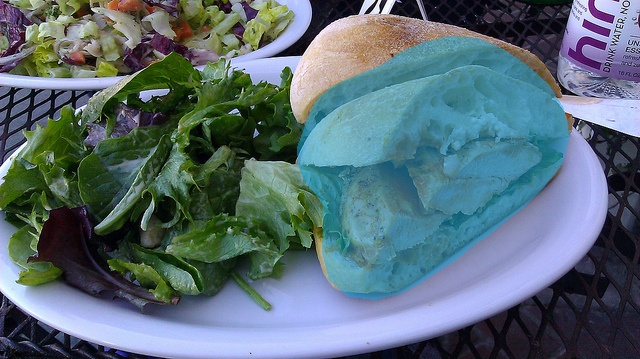}}
    {\includegraphics[height=0.62in]
    {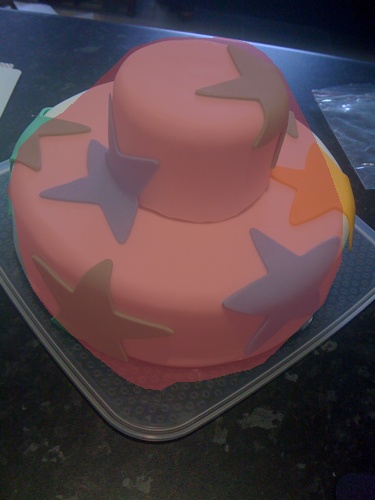}}
    {\includegraphics[height=0.62in]
    {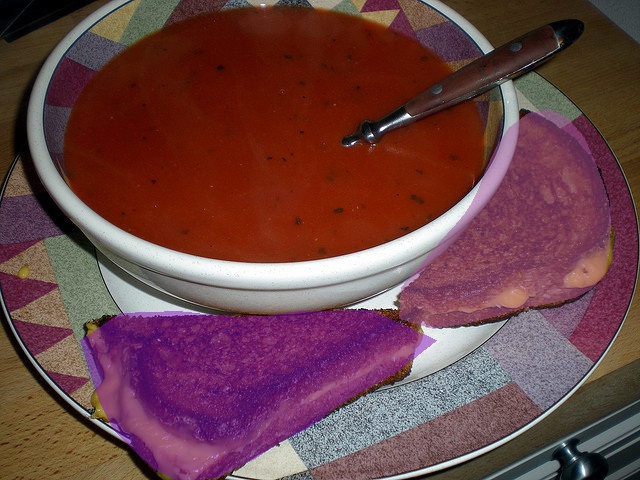}} \\
    \vspace{0.05in}
    {\includegraphics[height=0.62in]{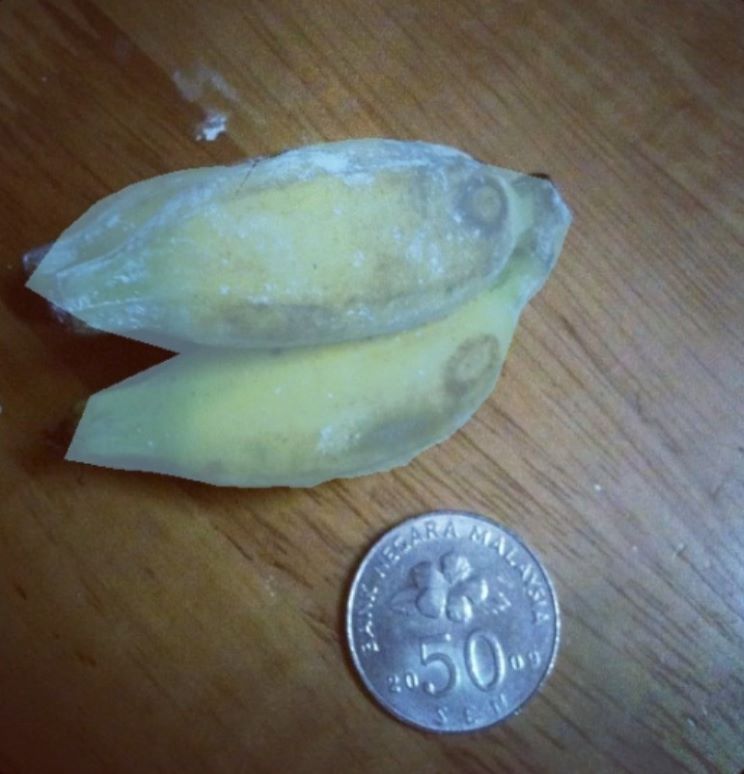}}
    {\includegraphics[height=0.62in]{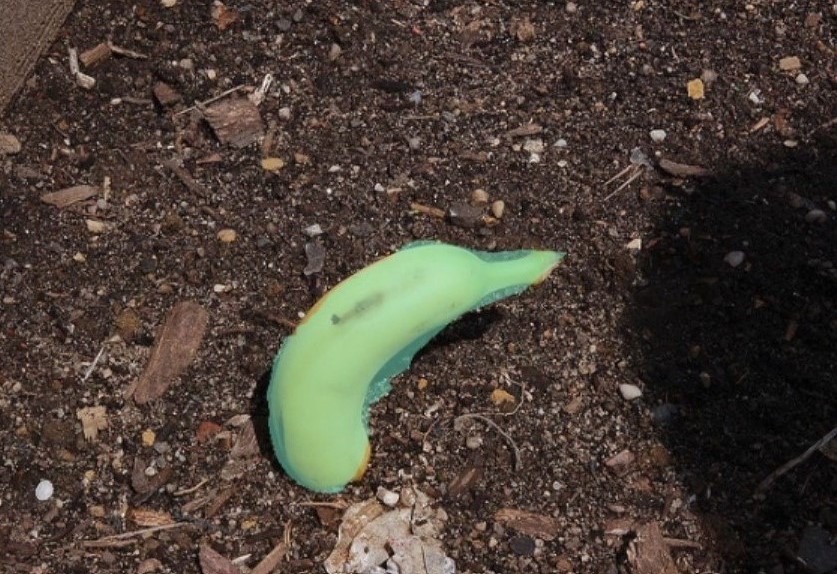}}
    {\includegraphics[height=0.62in]{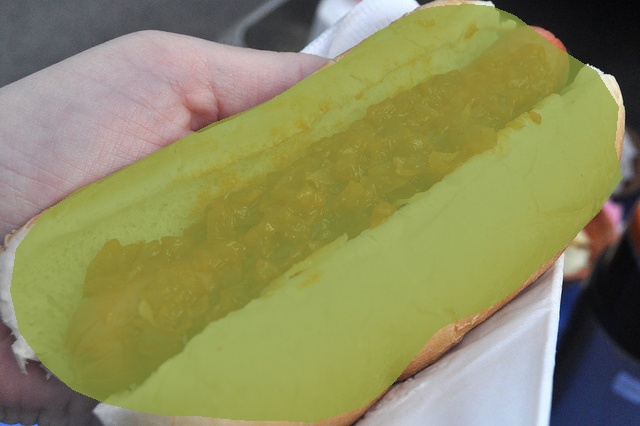}}
    {\includegraphics[height=0.62in]{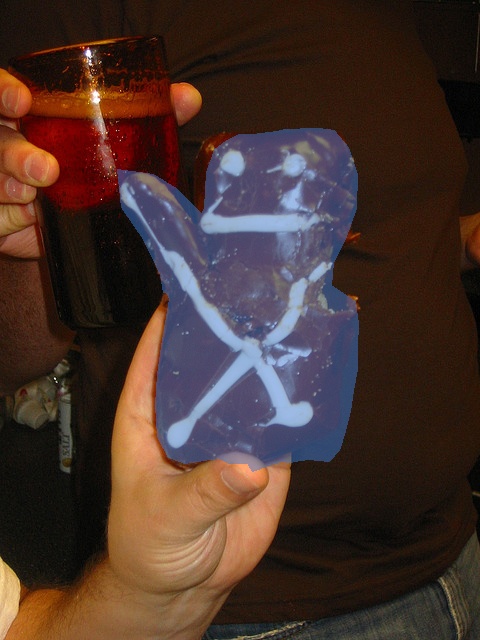}}
    {\includegraphics[height=0.62in]{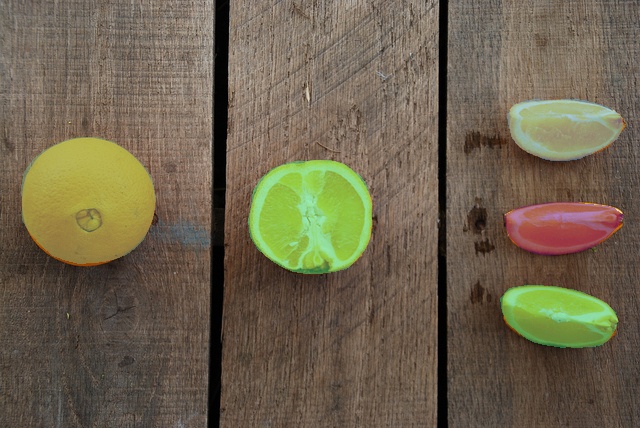}}
    {\includegraphics[height=0.62in]{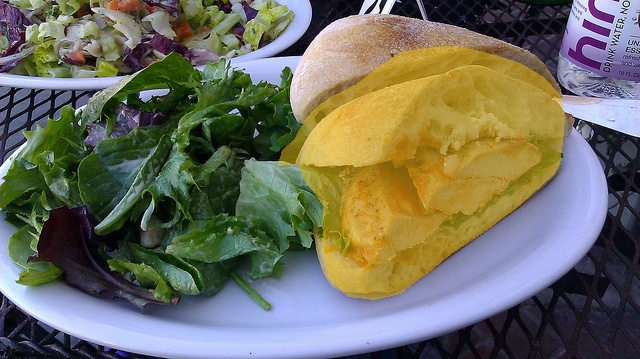}}
    {\includegraphics[height=0.62in]{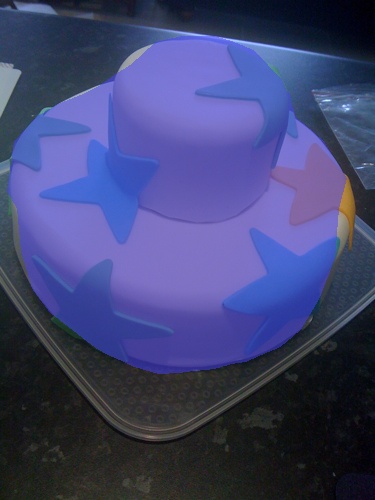}}
    {\includegraphics[height=0.62in]{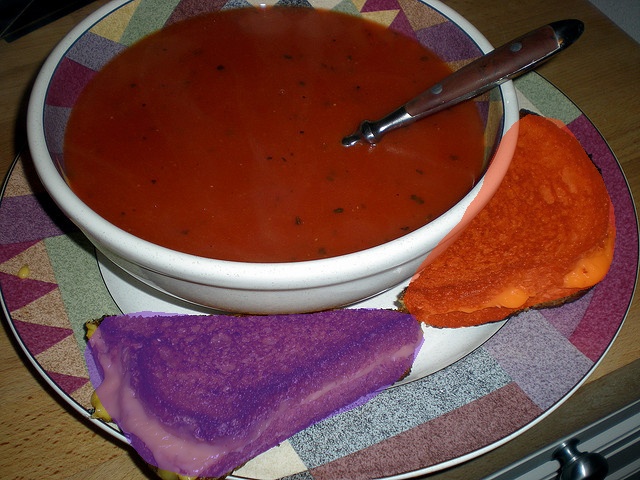}}
    \caption{Comparison of segmentation results of food images by PolarMask (upper part) and MP-PolarMask (lower part).}
    \label{Fig.main7} 
    \end{figure*}

\section{Experiment Results} 
\label{sec4}

We have conducted extensive comparisons with a number of state-of-the-arts on the COCO dataset (dataset A) and in particular the food images in the COCO dataset (dataset B). Our evaluation criteria encompass various essential image processing metrics: AP (Average Precision), AP$_{50}$ (AP at IoU $0.5$), AP$_{75}$ (AP at IoU $0.75$), AP$_L$ (AP for large objects), AP$_M$ (AP for medium-sized objects), AP$_S$ (AP for small objects), and the speed factor FPS (Frames Per Second). All the results are reported in \cref{tab2}.

\subsection{Validation on General Objects} \label{sec4.1}

In our experiment, except for ExtremeNet, which uses Hourglass-104 as its backbone, the backbones of all other models are based on ResNet-101. Based on PolarMask(side $=600$, 36 rays) and MP-PolarMask(side $=600$, 36 rays), we adjust the shorter side of test images to 600 pixels so as to enhance processing speed. To enrich our experiments and ensure fairness, we also test two backbones for MP-PolarMask: BFP \cite{7780460} and DCN \cite{9009086} (refer to the test in PolarMask++ \cite{9159936}). Regarding FPS, MP-PolarMask runs at $13.3$ FPS with the ResNet-101 backbone, slightly behind PolarMask's $13.9$ FPS. However, MP-PolarMask's AP of 35.5 surpasses PolarMask's AP of $32.1$. Remarkably, MP-PolarMask exhibits a noincrease in AP with only a slight decrease in FPS, highlighting its efficiency and competitive edge compared to PolarMask. If we adjust images' side size by using MP-PolarMask(side $=600$, 36 rays), apart from Tensormask, MP-PolarMask exhibits better performance in terms of AP, AP$_{50}$, AP$_{75}$, AP$_S$, AP$_M$, and AP$_{L}$ compared to all other models.
It is particularly noteworthy that, when we use DCN as the backbone for MP-PolarMask, a slight improvement is observed in AP, AP$_{50}$, AP$_{75}$, AP$_S$, AP$_M$, and AP$_L$ compared to Tensormask. Additionally, the FPS ratio to Tensormask is $9.2:2.1$.

\subsection{Validation on Food Objects} 
\label{sec4.2}

Food images are more challenging due to their irregular and concave shapes. From our evaluations, MP-PolarMask demonstrates superior performance compared to PolarMask in terms of AP, AP$_{50}$, AP$_{75}$, AP$_S$, AP$_M$, and AP$_L$. For example, MP-PolarMask improves AP by $\frac{(34.1-31.8)}{31.8} = 7.23\%$  and AP$_L$ by $\frac{(49.8-43.8)}{43.8} = 13.69\%$. If we switch the backbone of MP-PolarMask to DCN, there is a decrease of $4.8$ FPS compared to PolarMask, but the improvements of AP and AP$_L$ enlarge to $\frac{(35.8 -31.8)}{31.8} = 12.57\%$  and  $\frac{(51.2-43.8)}{43.8} = 16.89\%$, respectively.

In \cref{Fig.main7}, we show some instance segmentation results for food-related images generated by PolarMask and MP-PolarMask. There is a clear advantage of using MP-PolarMask, especially when objects are of concave shapes.

\section{Conclusions} \label{sec5}
As instant segmentation is a fundamental issue in computer vision, it is critical to perform the task in a real-time manner to facilitate downstream tasks. We propose a way to extend PolarMask to multiple Polar systems, thus achieving finer segmentation results. Through validation on the COCO dataset, MP-PolarMask demonstrates excellence in handling concave objects. However, food objects are still very challenging for the segmentation task as we do observe lower AP when comparing to the AP of general objects by MP-PolarMask. Future work may be directed to choosing more flexible auxiliary points and developing a better mask assembly method. 

{\small
\bibliographystyle{ieee_fullname}
\bibliography{ref}

\begin{thebibliography}{10}\itemsep=-1pt

\bibitem{aslan2020benchmarking}
Sinem Aslan, Gianluigi Ciocca, Davide Mazzini, and Raimondo Schettini.
\newblock Benchmarking algorithms for food localization and semantic segmentation.
\newblock {\em International Journal of Machine Learning and Cybernetics}, 11(12):2827--2847, 2020.

\bibitem{bolya2019yolact}
Daniel Bolya, Chong Zhou, Fanyi Xiao, and Yong~Jae Lee.
\newblock {YOLACT}: Real-time instance segmentation, 2019.

\bibitem{9159935}
Daniel Bolya, Chong Zhou, Fanyi Xiao, and Jae Lee~Yong.
\newblock {YOLACT++}: Better real-time instance segmentation.
\newblock {\em IEEE Transactions on Pattern Analysis and Machine Intelligence}, 44(2):1108--1121, 2022.

\bibitem{9009086}
Jiale Cao, Yanwei Pang, Jungong Han, and Xuelong Li.
\newblock Hierarchical shot detector.
\newblock In {\em 2019 IEEE/CVF International Conference on Computer Vision (ICCV)}, pages 9704--9713, 2019.

\bibitem{9010024}
Xinlei Chen, Ross Girshick, Kaiming He, and Piotr Dollar.
\newblock Tensormask: A foundation for dense object segmentation.
\newblock In {\em 2019 IEEE/CVF International Conference on Computer Vision (ICCV)}, pages 2061--2069, 2019.

\bibitem{9878483}
Bowen Cheng, Ishan Misra, Alexander~G. Schwing, Alexander Kirillov, and Rohit Girdhar.
\newblock Masked-attention mask transformer for universal image segmentation.
\newblock In {\em 2022 IEEE/CVF Conference on Computer Vision and Pattern Recognition (CVPR)}, pages 1280--1289, 2022.

\bibitem{cheng2022sparse}
Tianheng Cheng, Xinggang Wang, Shaoyu Chen, Wenqiang Zhang, Qian Zhang, Chang Huang, Zhaoxiang Zhang, and Wenyu Liu.
\newblock Sparse instance activation for real-time instance segmentation, 2022.

\bibitem{video-retrieval-PR}
Ting-Hui Chiang, Yi-Chun Tseng, and Yu-Chee Tseng.
\newblock A multi-embedding neural model for incident video retrieval.
\newblock {\em Pattern Recognition}, 130:108807, 2022.

\bibitem{Brabandere_2017_CVPR_Workshops}
Bert De~Brabandere, Davy Neven, and Luc Van~Gool.
\newblock Semantic instance segmentation for autonomous driving.
\newblock In {\em Proceedings of the IEEE Conference on Computer Vision and Pattern Recognition (CVPR) Workshops}, July 2017.

\bibitem{9010985}
Kaiwen Duan, Song Bai, Lingxi Xie, Honggang Qi, Qingming Huang, and Qi Tian.
\newblock Centernet: Keypoint triplets for object detection.
\newblock In {\em 2019 IEEE/CVF International Conference on Computer Vision (ICCV)}, pages 6568--6577, 2019.

\bibitem{9159936}
Xie Enze, Wang Wenhai, Ding Mingyu, Zhang Ruimao, and Luo Ping.
\newblock Polarmask++: Enhanced polar representation for single-shot instance segmentation and beyond.
\newblock {\em IEEE Transactions on Pattern Analysis and Machine Intelligence}, 44(9):5385--5400, 2021.

\bibitem{10.1007/978-3-319-10584-0_20}
Bharath Hariharan, Pablo Arbel{\'a}ez, Ross Girshick, and Jitendra Malik.
\newblock Simultaneous detection and segmentation.
\newblock In David Fleet, Tomas Pajdla, Bernt Schiele, and Tinne Tuytelaars, editors, {\em Proceedings of the Computer Vision -- ECCV 2014}, pages 297--312, Cham, 2014. Springer International Publishing.

\bibitem{8237584}
Kaiming He, Georgia Gkioxari, Piotr Dollár, and Ross Girshick.
\newblock Mask {R-CNN}.
\newblock In {\em 2017 IEEE International Conference on Computer Vision (ICCV)}, pages 2980--2988, 2017.

\bibitem{he2018mask}
Kaiming He, Georgia Gkioxari, Piotr Dollár, and Ross Girshick.
\newblock Mask {R-CNN}, 2018.

\bibitem{7780459}
Kaiming He, Xiangyu Zhang, Shaoqing Ren, and Jian Sun.
\newblock Deep residual learning for image recognition.
\newblock In {\em 2016 IEEE Conference on Computer Vision and Pattern Recognition (CVPR)}, pages 770--778, 2016.

\bibitem{Huang_2019_CVPR}
Zhaojin Huang, Lichao Huang, Yongchao Gong, Chang Huang, and Xinggang Wang.
\newblock Mask scoring {R-CNN}.
\newblock In {\em Proceedings of the IEEE/CVF Conference on Computer Vision and Pattern Recognition (CVPR)}, June 2019.

\bibitem{7780460}
Pang Jiangmiao, Chen Kai, Shi Jianping, Feng Huajun, Ouyang Wanli, and Lin Dahua.
\newblock Libra r-cnn: Towards balanced learning for object detection.
\newblock In {\em 2019 IEEE/CVF Conference on Computer Vision and Pattern Recognition (CVPR)}, 2019.

\bibitem{Law2020}
Hei Law and Jia Deng.
\newblock Cornernet: Detecting objects as paired keypoints.
\newblock In {\em Computer Vision – ECCV 2018: 15th European Conference}, volume 128, page 765–781, 2018.

\bibitem{8099955}
Y. Li, H. Qi, J. Dai, X. Ji, and Y. Wei.
\newblock Fully convolutional instance-aware semantic segmentation.
\newblock In {\em 2017 IEEE Conference on Computer Vision and Pattern Recognition (CVPR)}, pages 4438--4446, Los Alamitos, CA, USA, jul 2017. Proceedings of the IEEE Computer Society.

\bibitem{lin2018focal}
Tsung-Yi Lin, Priya Goyal, Ross Girshick, Kaiming He, and Piotr Dollár.
\newblock Focal loss for dense object detection, 2018.

\bibitem{10.1007/978-3-319-10602-1_48}
Tsung-Yi Lin, Michael Maire, Serge Belongie, James Hays, Pietro Perona, Deva Ramanan, Piotr Doll{\'a}r, and C.~Lawrence Zitnick.
\newblock Microsoft {COCO}: Common objects in context.
\newblock In David Fleet, Tomas Pajdla, Bernt Schiele, and Tinne Tuytelaars, editors, {\em Computer Vision -- ECCV 2014}, pages 740--755, Cham, 2014. Springer International Publishing.

\bibitem{8237640}
Shu Liu, Jiaya Jia, Sanja Fidler, and Raquel Urtasun.
\newblock Sgn: Sequential grouping networks for instance segmentation.
\newblock In {\em Proceedings of the 2017 IEEE International Conference on Computer Vision (ICCV)}, pages 3516--3524, 2017.

\bibitem{boundary}
Yu-Ting Liu, Jen-Jee Chen, Yu-Chee Tseng, and Frank~Y Li.
\newblock An auto-encoder multi-task lstm model for boundary localization.
\newblock {\em IEEE Sensors Journal}, 22(11):10940--10953, 2022.

\bibitem{10.5555/2969442.2969462}
Pedro~O. Pinheiro, Ronan Collobert, and Piotr Doll\'{a}r.
\newblock Learning to segment object candidates.
\newblock In {\em Proceedings of the 28th International Conference on Neural Information Processing Systems - Volume 2}, NIPS'15, page 1990–1998, Cambridge, MA, USA, 2015. MIT Press.

\bibitem{10.1007/978-3-319-46448-0_5}
Pedro~O. Pinheiro, Tsung-Yi Lin, Ronan Collobert, and Piotr Doll{\'a}r.
\newblock Learning to refine object segments.
\newblock In Bastian Leibe, Jiri Matas, Nicu Sebe, and Max Welling, editors, {\em Computer Vision -- ECCV 2016}, pages 75--91, Cham, 2016. Springer International Publishing.

\bibitem{poply2021refined}
Parth Poply and J~Angel~Arul Jothi.
\newblock Refined image segmentation for calorie estimation of multiple-dish food items.
\newblock In {\em 2021 International Conference on Computing, Communication, and Intelligent Systems (ICCCIS)}, pages 682--687. IEEE, 2021.

\bibitem{rahman2024kitchen}
Raiyan Rahman, Mohsena Chowdhury, Yueyang Tang, Huayi Gao, George Yin, and Guanghui Wang.
\newblock Kitchen food waste image segmentation and classification for compost nutrients estimation.
\newblock {\em arXiv preprint arXiv:2401.15175}, 2024.

\bibitem{tian2019fcos}
Zhi Tian, Chunhua Shen, Hao Chen, and Tong He.
\newblock {FCOS}: Fully convolutional one-stage object detection, 2019.

\bibitem{5G-CV}
Yu-Yun Tseng, Po-Min Hsu, Jen-Jee Chen, and Yu-Chee Tseng.
\newblock Computer vision-assisted instant alerts in {5G}.
\newblock In {\em 2020 29th International Conference on Computer Communications and Networks (ICCCN)}, 2020.

\bibitem{video-retrieval-CVPR}
Chia-Hui Wang, Yu-Chee Tseng, Ting-Hui Chiang, and Yan-Ann Chen.
\newblock Learning multi-scale representations with single-stream network for video retrieval.
\newblock In {\em Proceedings of the IEEE/CVF Conference on Computer Vision and Pattern Recognition (CVPR) Workshops}, pages 6166--6176, June 2023.

\bibitem{10.1007/978-3-030-58523-5_38}
Xinlong Wang, Tao Kong, Chunhua Shen, Yuning Jiang, and Lei Li.
\newblock {SOLO}: Segmenting objects by locations.
\newblock In Andrea Vedaldi, Horst Bischof, Thomas Brox, and Jan-Michael Frahm, editors, {\em Proceedings of the Computer Vision -- ECCV 2020}, pages 649--665, Cham, 2020. Springer International Publishing.

\bibitem{9157078}
Enze Xie, Peize Sun, Xiaoge Song, Wenhai Wang, Xuebo Liu, Ding Liang, Chunhua Shen, and Ping Luo.
\newblock Polarmask: Single shot instance segmentation with polar representation.
\newblock In {\em Proceedings of the 2020 IEEE/CVF Conference on Computer Vision and Pattern Recognition (CVPR)}, pages 12190--12199, 2020.

\bibitem{9746109}
Yuxuan Zhang and Wei Yang.
\newblock {BSOLO}: Boundary-aware one-stage instance segmentation {SOLO}.
\newblock In {\em Proceedings of the 2022 IEEE International Conference on Acoustics, Speech and Signal Processing (ICASSP 2022)}, pages 2594--2598, 2022.

\bibitem{8954244}
Xingyi Zhou, Jiacheng Zhuo, and Philipp P.~Kr\"ahenb\"uhl.
\newblock Bottom-up object detection by grouping extreme and center points.
\newblock In {\em 2019 IEEE/CVF Conference on Computer Vision and Pattern Recognition (CVPR)}, pages 850--859, Los Alamitos, CA, USA, jun 2019. IEEE Computer Society.

\end{thebibliography}
}
\end{document}